\documentclass{article}

\usepackage[preprint]{neurips_2026}

\usepackage[utf8]{inputenc} 
\usepackage[T1]{fontenc}    
\usepackage{hyperref}       
\usepackage{url}            
\usepackage{booktabs}       
\usepackage{amsfonts}       
\usepackage{nicefrac}       
\usepackage{microtype}      
\usepackage{xcolor}         
\usepackage{amsmath}
\usepackage{amssymb}
\usepackage{mathtools}
\usepackage{amsthm}
\theoremstyle{plain}
\newtheorem{theorem}{Theorem}[section]

\newtheorem{lem}[theorem]{Lemma}

\theoremstyle{definition}
\newtheorem{defi}[theorem]{Definition}
\newtheorem{ass}[theorem]{Assumption}
\theoremstyle{remark}
\newtheorem{remark}[theorem]{Remark}
\usepackage{microtype}
\usepackage{graphicx}
\RequirePackage{algorithm}
\RequirePackage{algorithmic}
\usepackage{booktabs} 
\usepackage{xcolor}
\newcommand{\Norm}[1]{\left\|#1\right\|}
\usepackage{multirow} 
\usepackage{makecell} 

\newcommand{\LineComment}[1]{\hfill$\rhd\ $\text{#1}}
\def \E {\mathbb{E}}
\def \R {\mathbb{R}}

\def \e {\mathbf{e}}

\def \g {\mathbf{g}}

\def \s {\mathbf{s}}
\def \p {\mathbf{p}}
\def \rr {\mathbf{r}}

\def \z {\mathbf{z}}

\def \vv {\mathbf{v}}
\def \w {\mathbf{w}}
\def \x {\mathbf{x}}
\def \y {\mathbf{y}}
\def \A {\mathcal{A}}

\def \W {\mathcal{W}}

\def \C {\mathcal{C}}

\newcommand{\EC}[1]{\mathbb{E}_{\mathcal{C}}\left[#1\right]}
\usepackage[capitalize,noabbrev]{cleveref}
\hypersetup{
    colorlinks,
    breaklinks,
    urlcolor = black,
    linkcolor = blue,
    citecolor = blue,
}

\title{Distributed Online Convex Optimization with Compressed Communication: Optimal Regret and Applications}

\author{
    Sifan Yang\textsuperscript{\rm 1,2},~~Dan-Yue Li\textsuperscript{\rm 1,2},~~Lijun Zhang\textsuperscript{\rm 1,2}\\
    \textsuperscript{\rm 1}National Key Laboratory for Novel Software Technology, Nanjing University, Nanjing, China \\
    \textsuperscript{\rm 2}School of Artificial Intelligence, Nanjing University, Nanjing, China \\
    \texttt{\{yangsf, lidy, zhanglj\}@lamda.nju.edu.cn}
}
\begin{document}

\maketitle

\begin{abstract}
Distributed online convex optimization (D-OCO) is a powerful paradigm for modeling distributed scenarios with streaming data.  However, the communication cost between local learners and the central server is substantial in large-scale applications. To alleviate this bottleneck, we initiate the study of  D-OCO with compressed communication. Firstly, to quantify the compression impact, we establish the $\Omega(\delta^{-1/2}\sqrt{T})$ and $\Omega(\delta^{-1}\log{T})$ lower bounds for convex and strongly convex loss functions, respectively, where $\delta \in (0,1]$ is the compression ratio. 
Secondly, we propose an optimal algorithm, which enjoys regret bounds of $O(\delta^{-1/2}\sqrt{T})$ and $O(\delta^{-1} \log T)$ for convex and strongly convex loss functions, respectively.
Our method incorporates the  \emph{error feedback mechanism} into the \emph{Follow-the-Regularized-Leader} framework to address the coupling between the compression error and the projection error. Furthermore, we employ the \emph{online compression strategy} to mitigate the accumulated error arising from the bidirectional compression. Our online method has great generality, and can be extended to the offline stochastic setting via online-to-batch conversion.
 We establish convergence rates of $O(\delta^{-1/2}T^{-1/2})$ and $O(\delta^{-1} T^{-1})$ for convex and strongly convex loss functions, respectively, providing the \emph{first}  guarantees for distributed non-smooth optimization with compressed communication and domain constraints.
\end{abstract}

\section{Introduction}
Distributed online convex optimization (D-OCO) \citep{patel2023federated} is a fundamental paradigm for modeling distributed sequential decision-making problems, such as mobile keyboard predictions \citep{hard2018federated,chen2019federated}, self-driving vehicles \citep{elbir2020federated}, recommendation systems \citep{shi2021federated,liang2021fedrec++,khan2021payload}, among others. Specifically, it is formulated as a repeated game between an adversary and  a set of $n$ local learners connected to a central server. At each round $t\in[T]$, each online learner plays a \emph{global decision} $\w^t$ from a convex set $\W\subseteq\R^d$. After the decision $\w^t$ is committed, learner $i$ suffers a local loss $f^t_i(\w^t)$, where $f^t_i(\cdot): \W\to\R$ is a convex function. Subsequently, learner $i$ sends its local information to the central server, which aggregates the received data and broadcasts the global information back to all learners. Upon receiving this message, local learners synchronously update the decision $\w^{t+1}$. To measure the performance of learners, we select the \emph{standard regret}:
\begin{equation}\label{gr}
    R_T=\frac{1}{n}\sum_{t=1}^T\sum_{i=1}^n\left(f_i^t(\w^t)- f_i^t(\w^\star)\right),
\end{equation}
where $\w^\star=\arg\min_{\w\in\W}\frac{1}{n}\sum_{t=1}^T\sum_{i=1}^nf_i^t(\w)$ is the best decision chosen in hindsight.

It is straightforward to verify that if each learner transmits its local gradient to the central server and updates the decision with the aggregated gradient from the server, the distributed setting  reduces to the single-node case, and the procedure attains the optimal regret bound for single-node OCO \citep{shalev2012online,hazan2016introduction}. In practice, however, transmitting full gradients between the learners and the server may incur substantial \emph{communication overhead},  limiting the applicability of this approach in large-scale distributed problems.  To alleviate this bottleneck in D-OCO, we conduct the \emph{first} investigation of communication-efficient D-OCO, where learners and the server exchange compressed messages. We  note that prior works have explored compressed communication in decentralized online convex optimization \citep{tu2022distributed,CAO2023111186,Yang2026EfficientDOCO}, but their methods are tailored for peer-to-peer topologies. Consequently, they are incompatible with our distributed setting due to the fundamental differences in network architecture.

Naively using the compressed gradients, e.g., updating with the sign of the gradients, may result in the unbounded accumulation of the compression error, which prevents convergence even in the offline optimization \citep{karimireddy2019error}. To address this issue, the error feedback (EF) mechanism \citep{seide20141} is widely adopted to control the compression error in unconstrained settings \citep{alistarh2018convergence,karimireddy2019error,stich2020error,richtarik2021ef21,huang2022lower}. At each round, EF records the accumulated error of the past compression steps and compresses the sum of the error and gradient, effectively bounding the total compression error over time.  However, a critical challenge arises when extending the existing EF-based optimization methods to the distributed settings with domain constraints. Since each learner must project its decision onto the feasible domain per update, it results in an additional projection error, which \emph{couples} with the compression error and leads to unbounded error accumulation. Unlike unconstrained optimization where EF alone suffices to control the compression error, the coupled error in constrained settings necessitates novel techniques for effective handling.

In this paper, we first develop  Distributed Follow-the-Compressed-Leader (D-FTCL), which applies EF on both learner and server sides to enable bidirectional compressed communication. More importantly, D-FTCL adapts EF to the Follow-the-Regularized-Leader (FTRL) framework \citep{hazan2016introduction,orabona2019modern}, which ensures that the projection error does not propagate to subsequent compression steps, thereby decoupling the projection error from the compression error.  D-FTCL enjoys regret bounds of $O(\delta^{-1}\sqrt{T})$ and $O(\delta^{-2} \log T)$ for convex and strongly convex loss functions, respectively, where $\delta \in (0,1]$ denotes the compression ratio that characterizes the quality of compression and $\delta=1$ means no compression. 
Furthermore, to quantify the impact of compression, we establish the $\Omega(\delta^{-1/2}\sqrt{T})$ and $\Omega(\delta^{-1}\log T)$ lower bounds for convex and strongly convex loss functions, respectively.  Thus, while D-FTCL successfully achieves efficient communication, there exists a gap between its regret guarantees and the lower bounds. This prompts a natural question: \emph{is it possible  to further improve the regret bound of D-FTCL?}

Through careful analysis, we find that the bidirectional compression in D-FTCL introduces the amplified compression error, which results in suboptimal regret bounds. Motivated by previous work \citep{huang2022lower,Yang2026EfficientDOCO}, we employ the online compression strategy  to reduce the compression error, which recursively compresses the \emph{residual} for multiple rounds. However, since the online compression strategy leverages the blocking update mechanism to ensure the communication round remains $1$ per update, the communication between the local learners and the server becomes \emph{asynchronous} due to transmission latency. To tackle this issue,  we rigorously characterize the deviation caused by the delayed updates in the analysis.  Our method, named Distributed Follow-the-Fast-Compressed-Leader (D-FTFCL), enhances the regret bounds to $O(\delta^{-1/2}\sqrt{T})$ and $O(\delta^{-1} \log T)$ for convex and strongly convex loss functions, respectively, thereby achieving optimal regret bounds for D-OCO with compressed communication.

 Our online methods have \emph{broad applicability}, and thus can be extended to the distributed offline stochastic convex optimization with domain constraints. Previous work on distributed offline optimization with compressed communication typically assumes either smooth loss functions or an unconstrained optimization domain \citep{karimireddy2019error,richtarik2021ef21,huang2022lower,Gao2024EControl,islamov2025safeef}. Consequently, the problem of distributed convex non-smooth optimization under domain constraints remains \emph{unsolved}. To bridge this gap, we  develop a distributed variant of anytime online-to-batch conversion  \citep{cutkosky2019anytime} and integrate our online method into this framework. Our proposed algorithm obtains a  last-iterate convergence rate of $O(\delta^{-1/2} T^{-1/2})$ for convex loss functions, matching the lower bound established by \citet{islamov2025safeef}. For strongly convex loss functions, we establish a faster rate of  $O\left(\delta^{-1} T^{-1}\right)$, which is optimal with respect to $T$ \citep{Agarwal2012OracleLowerBound}.  To the best of our knowledge, these are the  \emph{first} guarantees for distributed convex non-smooth optimization with compressed communication and domain constraints.


\section{Related Work}

In this section, we briefly review the related work on distributed offline optimization and distributed online convex optimization.

\subsection{Distributed Offline Optimization}\label{doo}
Distributed offline optimization has attracted significant attention in the past decades due to its wide applicability in distributed machine learning tasks \citep{nedic2009distributed,dean2012large,yuan2016convergence,tang2018communication,swenson2022distributed}.   Although prior distributed algorithms have established solid theoretical guarantees, the massive information exchange between local learners and the central server may result in a critical communication bottleneck, particularly in high-dimensional and large-scale scenarios.

To alleviate communication overhead, extensive research has integrated compression mechanisms into distributed offline optimization algorithms \citep{seide20141,stich2018sparsified,bernstein2018signsgd,karimireddy2019error,huang2022lower,islamov2025safeef}. Rather than broadcasting the full vector $\mathbf{x} \in \mathbb{R}^d$, they transmit compressed data $\mathcal{C}(\mathbf{x})$, where $\mathcal{C}(\cdot): \mathbb{R}^d\to\mathbb{R}^d$ is an operator designed for efficient transmission. A quintessential example is the sign operator, which outputs the sign of each coordinate, thereby significantly reducing the communication cost.

To mitigate errors induced by the sign compressor, \citet{seide20141} pioneer the error feedback (EF) mechanism, which tracks the past compression error and compresses the sum of the error and the gradient, though they do not provide theoretical guarantees. Subsequently, \citet{karimireddy2019error} provide a unified theoretical framework for EF with general compressors. However, their analysis is limited to unidirectional compression. To achieve bidirectional compression, \citet{Stich2020CommunicationCompression} and \citet{beznosikov2023biased} provide the analysis for EF under the distributed setting. In recent years, \citet{richtarik2021ef21} introduce  a novel error feedback technique, termed EF21, which compresses the residual of the transmitted data and the gradient.  To further mitigate the compression error, \citet{huang2022lower} design the fast compressed communication (FCC), which recursively applies the standard compressor for multiple rounds.  By utilizing FCC, they establish the nearly optimal convergence rates for non-convex smooth loss functions.

 Despite the extensive research on incorporating EF into the offline optimization, existing methods typically require the loss function to be smooth or the feasible domain to be unconstrained  \citep{seide20141,stich2018sparsified,karimireddy2019error,richtarik2021ef21,huang2022lower}. The only work that addresses offline distributed convex non-smooth optimization is \citet{islamov2025safeef}, which investigates the safe constraint setting. They establish a convergence rate of $O(\delta^{-1}T^{-1/2})$, and a lower bound of $\Omega(\delta^{-1/2}T^{-1/2})$ for unconstrained optimization. However, their method relies on the specific structure of safe constraints and fails to cover domain constraints.   Although \citet{islamov2025safeef} further provide an analysis for EF21 in the convex smooth setting with domain constraints, they assume smooth loss functions and are restricted to unidirectional compression.  It still remains an open problem whether EF is effective for convex non-smooth optimization with domain constraints.

\subsection{Distributed Online Convex Optimization}

To address the distributed optimization with streaming data,  \citet{patel2023federated} investigate distributed online convex optimization (D-OCO), where the loss function is time-varying and the learners have to continuously update decisions to minimize the cumulative regret. They consider a different regret defined as $R_T^\prime=\frac{1}{n}\sum_{i=1}^n\left(\sum_{t=1}^Tf_i^t(\w^t_i)-f_i^t(\w^\star)\right),$
where $\w^\star=\arg\min_{\w\in\W}\frac{1}{n}\sum_{t=1}^T\sum_{i=1}^nf_i^t(\w)$. $R_T^\prime$ measures the performance of each learner by using the local loss functions. As noted by \citet{patel2023federated}, communication offers no benefit  because learners can independently execute OGD \citep{zinkevich2003online} to achieve the optimal regret bound without any information exchange. In contrast, the regret defined in \eqref{gr} evaluates the performance of the decision against the global loss. This regret is more realistic for distributed learning tasks, where the ultimate goal is to train a consensus model. Moreover, when loss functions are fixed over rounds, our  regret coincides with the offline optimization objective.
Furthermore, they pivot to  the  bandit feedback setting, where the local learners can only access values of the loss functions \citep{agarwal2010optimal,shamir2017optimal}. In this regime, they demonstrate that collaboration helps to achieve tighter regret bounds.  

We also notice a parallel line of research investigating decentralized online convex optimization, where local learners interact over a network modeled by an undirected graph \citep{yan2012distributed,hosseini2013online,zhang2017projection,wan2020projection,wan2022projection,wang2023distributed,wan2024nearly,JMLR:2025:Wan,wan2025black}. To reduce the communication cost among learners, recent efforts have further explored compressed communication within this setting \citep{tu2022distributed,CAO2023111186,Yang2026EfficientDOCO}. However, in decentralized settings, each learner maintains an individual local decision and communicates in a peer-to-peer scheme without a central server for aggregation. This architecture fundamentally differs from the distributed setting we consider. Consequently, algorithms tailored for peer-to-peer topologies cannot be directly extended to our setting due to intrinsic differences in communication protocols and network structures.

\section{Preliminaries}
In this section, we  present necessary preliminaries including assumptions, definitions, and techniques used in this paper.

Similar to previous studies on single-node OCO, we introduce the following standard assumptions \citep{hazan2016introduction,orabona2019modern}.

\begin{ass}[convexity]\label{convex}
    The loss function $f_{i}^t (\cdot)$ of each learner $i \in [n]$ is convex over the feasible domain $\W$.
\end{ass}
\begin{ass}[strong convexity] \label{sconvex}
    The loss function $f_{i}^t (\cdot)$ of each learner $i \in [n]$ is $\mu$-strongly convex over the domain $\W$,  i.e., it holds that $f_{i}^t(\mathbf{y}) \ge f_{i}^t(\mathbf{x}) + \langle \nabla f_{i}^t(\mathbf{x}), \mathbf{y} - \mathbf{x} \rangle + \frac{\mu}{2} \|\mathbf{y} - \mathbf{x}\|^{2}$, for $\forall \x,\y \in \W$.
\end{ass}
\begin{ass}[bounded gradient norm] \label{gradient}
    The gradient of function $f_{i}^t(\cdot)$ of each learner $i \in [n]$ is bounded by $G$ over the domain $\W$, i.e., it holds that $\Norm{\nabla f_{i}^t(\w)}\leq G$, for $\forall  \w \in \W$ and $\Norm{\frac{1}{n}\sum_{i=1}^n\nabla f_{i}^t(\w)}\leq G$, for $\forall  \w \in \W$
\end{ass}
\begin{ass}[bounded domain]\label{domain}
    The convex set $\W$ contains the origin $\textbf{0}$, i.e., $\textbf{0} \in \W$, and it is bounded by $D$, i.e., it holds that $\Norm{\x-\y} \leq D$, for $\forall \x,\y \in \W$.
\end{ass}

Next, we introduce a common assumption governing the communication process in our distributed setting.

\begin{ass}[communication protocol]
At each round, each learner $i$ and the central server are allowed to communicate with each other exactly once.
\end{ass}

 A compressor $\mathcal{C}(\cdot): \mathbb{R}^d \to \mathbb{R}^d$  is a mapping whose output can be efficiently transmitted.  In this paper, we consider a general class of compressors with the
following definition \citep{koloskova2019decentralized}.

\begin{defi}[compressor]\label{compress} 
A compressor $\mathcal{C}(\cdot): \mathbb{R}^d \to \mathbb{R}^d$ is called $\delta$-contractive if the following holds
\begin{equation*}
    \mathbb{E}_{\C}\left[\Norm{\C(\x)-\x}^2\right]\leq (1-\delta)\Norm{\x}^2, \quad \forall \x \in \mathbb{R}^d,
\end{equation*}
for some $\delta \in (0,1]$, where the expectation $\EC{\cdot}$ is taken over the randomness of the compressor.
\end{defi}

To mitigate the error of compressors, \citet{huang2022lower} design the \textit{fast compressed communication} (FCC), as summarized in  Algorithm~\ref{rpc}. The core idea is to recursively apply the standard compressor for $L$ rounds and transmit the compressed data at each round, which involves $L$ rounds of communication. When $L=1$, FCC degenerates to the standard compressor. 
We state the following lemma to provide the guarantee of FCC.
\begin{lem}[Lemma 2 in \citet{huang2022lower}] \label{lem:he} Given a $\delta$-contractive compressor $\C(\cdot)$ and for any compression rounds $L\geq1$,  Algorithm \ref{rpc} ensures 
\begin{equation*}
        \mathbb{E}_{\C}\left[\Norm{\mathbf{r}^{L+1}-\x}^2\right]\leq (1-\delta)^L\Norm{\x}^2, \forall \x \in \mathbb{R}^d.
    \end{equation*}
\end{lem}
 Lemma \ref{lem:he} demonstrates that the error in FCC decreases exponentially as $L$ grows, but such a result is achieved at the expense of $L$ communication rounds.. 

\begin{algorithm}[t]
   \caption{FCC($\x,\C(\cdot), L$, receiver)}
   \label{rpc}
\begin{algorithmic}[1]
    \STATE {\bfseries Input:} data $\x$, compressor $\C(\cdot)$,  compression round $L$, receiver
    \STATE Initialize $\mathbf{r}^1 = \textbf{0}$
     \FOR{$k=1$ to $L$}
     \STATE Send $\mathbf{c}^{k} = \C(\x-\mathbf{r}^k)$ to the receiver
     \STATE Calculate $\mathbf{r}^{k+1} = \mathbf{r}^k+\mathbf{c}^{k}$
     \ENDFOR
     \STATE \textbf{Return} $\mathbf{r}^{L+1}$
\end{algorithmic}
\end{algorithm}

\section{Main Results}
In this section, we first present an efficient algorithm for D-OCO with compressed communication, as well as the corresponding theoretical guarantees. Then, we derive the lower bounds for D-OCO with compressed communication. Finally, we develop an improved method that achieves the optimal regret bounds.
\subsection{Follow-the-Compressed-Leader}
 Before presenting our algorithms, we briefly highlight the key challenges in applying EF mechanism to optimization problems with constrained domains. In the single-node unconstrained setting, the learner initializes $\e^1 = \mathbf{0}$ to track compression errors across iterations and performs the update under the gradient descent framework:
\begin{equation*}\label{ef-e}
    \e^{t+1}=\e^t+\g^t-\C(\mathbf{e}^t+\g^t)
\end{equation*}
\begin{equation*}\label{ef:up}
    \w^{t+1}=\w^t-\eta \C(\mathbf{e}^t+\g^t),
\end{equation*}
where $\g^t$ is the gradient at round $t$ and $\eta$ is the learning rate. Let $\hat{\w}^{t+1}$ denote the sequence obtained using the exact gradient, i.e., $\hat{\w}^{t+1}=\hat{\w}^t-\eta \g^t$. EF ensures the deviation remains bounded:
\begin{equation}\label{uc:ef}
    \EC{\Norm{\w^t-\hat{\w}^t}}\leq\EC{\eta\Norm{\e^t}}\leq O(\delta^{-1}\eta).
\end{equation}
Consequently, by selecting an appropriate learning rate $\eta$, the compression error can be effectively controlled, ensuring that the algorithm converges despite the effect induced by compression. However, applying EF to D-OCO encounters a critical issue due to the \emph{coupling} of the compression error and the projection error. Specifically, since each learner has to project its decision onto the feasible domain, i.e.,
 \begin{equation}\label{project-update}
 \w^{t+1}=\Pi_{\W}\left[\w^t-\eta \C(\mathbf{e}^t+\g^t)\right],
 \end{equation}
it incurs an additional projection error and the projection error will propagate to all subsequent compression steps,  thereby invalidating the guarantee in \eqref{uc:ef}. This interaction between projection and compression causes the total error to become \emph{uncontrollable}.

 \begin{algorithm}[t]
\caption{Distributed Follow-the-Compressed-Leader (D-FTCL)}
\label{alg:1}
\begin{algorithmic}[1]
\REQUIRE learning rate $\eta$, compressor $\C(\cdot)$
\STATE Initialize  $\w^1 =\textbf{0}$,  $\hat{\e}^1 = \textbf{0}$ and $\e_i^1 = \textbf{0}, \forall i \in[n]$
\FOR{$t = 1, \ldots, T$}
\STATE \textbf{for each learner} $i = 1, \ldots, n$ \textbf{in parallel}
\STATE $\quad $ Play the decision $\w^t$ and receive gradient $\g_i^t$
\STATE $\quad $ Send $\mathbf{v}_i^t = \C(\e_i^t + \g_i^t)$ to server and compute $\e_i^{t+1} = \e_i^t + \g_i^t - \mathbf{v}_i^t$
\STATE \textbf{for server do}
\STATE $\quad $ Compute $\mathbf{v}^t = \frac{1}{n} \sum_{i=1}^n \mathbf{v}_i^t$ and send $\s^t = \C(\hat{\e}^t + \mathbf{v}^t)$ to all learners
 \STATE $\quad $ Compute  $\hat{\e}^{t+1} = \hat{\e}^t  + \mathbf{v}^t - \s^t$
  \STATE \textbf{for each learner} $i = 1, \ldots, n$ \textbf{in parallel}
\STATE $\quad $ Update the decision $\w^{t+1}$ according to \eqref{eq:ftcl-convex} or \eqref{eq:ftcl-sconvex} 
\ENDFOR
\end{algorithmic}
\end{algorithm}
 To address the aforementioned limitation, we develop D-FTCL by integrating EF into the Follow-The-Regularized-Leader (FTRL) framework \citep{hazan2007logarithmic,hazan2016introduction} to \emph{decouple} the projection error from the compression error, as summarized in Algorithm \ref{alg:1}. Additionally, we employ the EF mechanism on both learner and server sides to achieve bidirectional compression. 
 Specifically, at each round $t$, the learner $i$ plays the global decision $\w^t$ and observes the gradient $\g_i^t$. Then it broadcasts the compressed message $\mathbf{v}_i^t = \C(\e_i^t + \g_i^t)$ to the server and updates $ \e_i^{t+1}=\e_i^t+\g_i^t-\C(\mathbf{e}_i^t+\g_i^t)$. 
 Upon receiving all messages from the learners, the server computes the aggregate information $\vv^t=\frac{1}{n}\sum_{i=1}^n \vv_i^t$. Next, it transmits $\s^t = \C(\hat{\e}^t + \mathbf{v}^t)$ back to learners and updates $\hat{\e}^{t+1} = \hat{\e}^t  + \mathbf{v}^t - \s^t$. In contrast to previous methods that update within the gradient descent framework, each learner updates the decision as follows:
\begin{equation}\label{eq:ftcl-convex}
    \w^{t+1}=\arg\min_{\w\in\W}\left\langle\sum_{k=1}^t\s^{k},\w\right\rangle+\frac{1}{\eta}\Norm{\w}^2.
\end{equation} 
Since the decision $\w^{t+1}$ is updated with the historical compressed information, the projection error is confined to the current update step. Thus, we can prevent the error from accumulating across rounds, thereby successfully avoiding the coupling between the compression error and the projection error. 
If the loss function is $\mu$-strongly convex, each learner updates the decision according to:
\begin{equation}\label{eq:ftcl-sconvex}
\w^{t+1}=\arg\min_{\w\in\W}\left\langle\sum_{k=1}^t\s^{k},\w\right\rangle+\frac{\mu}{2}\sum_{k=1}^t \Norm{\w-\w^k}^2. 
\end{equation}

In the following, we establish the guarantees for D-FTCL.

\begin{theorem}\label{thm:1}
Under the Assumptions \ref{convex}, \ref{gradient} and \ref{domain}, by setting the learning rate $\eta=\frac{\delta D}{G\sqrt{T}}$, Algorithm \ref{alg:1} ensures 
\begin{equation*}
    \EC{R_T}\leq O(\delta^{-1}DG\sqrt{T}).
\end{equation*}
\end{theorem}

\begin{theorem}\label{thm:2}
Under the Assumptions \ref{sconvex}, \ref{gradient} and \ref{domain}, Algorithm \ref{alg:1} ensures 
\begin{equation*}
    \EC{R_T}\leq O\left(\mu^{-1}\delta^{-2}(G+\mu D)^2\log T\right).
\end{equation*}
\end{theorem}
\begin{remark}
Our algorithm achieves the $O(\delta^{-1}\sqrt{T})$ and $O\left(\delta^{-2} \log T\right)$ regret bounds for convex and $\mu$-strongly convex loss functions, respectively, which are the first theoretical guarantees for D-OCO with compressed communication and domain constraints.  
\end{remark}

\subsection{Lower Bounds}
To verify the optimality of our proposed method, we establish the lower bounds for D-OCO with compressed communication. To derive lower bounds for D-OCO with compressed communication, the key challenge is characterizing the effect of compression. Inspired by  \citet{huang2022lower}, we model the compressor as probabilistic communication failure, where transmitted information may not reach its destination at some rounds. We assume learners communicate exclusively with the central server and operate synchronously, ensuring all learners maintain identical decision. Since we focus on the online setting,  distinct from the construction in \citet{huang2022lower}, we adopt the randomized gossip compressor $\C(\cdot): \R^d\to \R^d$ \citep{koloskova2019decentralized}, which outputs $\C(\x)=\x$ with probability $\delta$ and $\C(\x)=\textbf{0}$ otherwise. Thus, each learner successfully transmits information to the server with probability $\delta$ per round, implying that the expected number of rounds for a successful transmission is $\lceil 1/\delta\rceil$. In this way, we can characterize the effect of compression by analyzing the regret degradation caused by the intermittent communication. We formally state the lower bounds in the following theorems.

\begin{theorem}\label{thm:7}
    Given the feasible domain $\W=[\frac{-D}{2\sqrt{d}},\frac{D}{2\sqrt{d}}]^d$ For any D-OCO algorithm, if  $T \geq (1-\delta)/\delta$, there exists a sequence of  convex loss functions satisfying Assumption~\ref{gradient} and a compressor satisfying Definition~\ref{compress} such that
    \begin{equation*}
        \EC{R_T} \geq \frac{DG\sqrt{T}}{2^3\sqrt{\delta}}.
\end{equation*}
\end{theorem}

\begin{theorem}\label{thm:8}
    Given the feasible domain $\W=[0,\frac{D}{\sqrt{d}}]^d$ For any D-OCO algorithm, if $T \geq (16+\delta)/ \delta $, there exists a sequence of  $\mu$-strongly convex loss functions satisfying Assumption~\ref{gradient} and a compressor satisfying Definition~\ref{compress} such that
    \begin{equation*}
        \EC{R_T} \geq \frac{\mu D^2(\log_{16}(15\delta T)-2)}{2^{18}  \delta }.
\end{equation*}
\end{theorem}
\begin{remark}
We establish $\Omega(\delta^{-1/2}\sqrt{T})$ and $\Omega(\delta^{-1}\log{T})$ lower bounds for convex and strongly convex loss functions, respectively. When there is no compression ($\delta=1$), our results recover the classical lower bounds for single-node OCO \citep{Abernethy2008Optimal,Hazan2014Beyond}.
\end{remark}

\begin{remark}
    One might notice that the dependence on the compression ratio $\delta$ of upper bounds in Theorems~\ref{thm:1} and \ref{thm:2} does not match the lower bounds.  Therefore, a natural question arises: \emph{is it possible to further improve the upper bounds?} We provide an affirmative answer by developing an improved algorithm in the next section.
\end{remark}

\subsection{Our Improved Algorithm}

Through careful analysis, we find that the suboptimal regret bounds are primarily caused by the bidirectional compression scheme. The server re-compresses the aggregated messages that have already been compressed by learners, leading to an amplification of the compression error. In fact, if we apply the \emph{unidirectional compression}, D-FTCL can obtain the optimal regret bounds with the detailed proof provided in Appendix~\ref{appendix:unidierectional}. Thus, to reduce the regret of D-FTCL, we need to control the  error caused by compression. In offline optimization,  \citet{huang2022lower} design the fast compressed communication (FCC) to reduce the compression error, which recursively compresses the residual over $L$ rounds. This multi-round approach is not allowed in the online setting, where learners must update their decisions instantaneously to cope with streaming data.
 \begin{algorithm}[t]
\caption{Distributed Follow-the-Fast-Compressed-Leader (D-FTFCL)}
\label{alg:2}
\begin{algorithmic}[1]
\REQUIRE  learning rate $\eta$, compressor $\C(\cdot)$, block size $L$
\STATE Initialize $\w^1 =\textbf{0}$, $\hat{\e}^1 = \textbf{0}$ and  $\e_i^1 = \textbf{0}, \forall i \in[n]$
\FOR{$b = 1, \ldots, T/L$}
\STATE \textbf{for each learner} $i = 1, \ldots, n$ \textbf{in parallel}
\STATE $\quad$ \textbf{for} $t = (b-1)L+1, \ldots, bL$ \textbf{do}
\STATE  $\quad$   $\quad$  Play the decision $\w^b$ and receive gradient $\g_i^t$
\STATE  $\quad$  \textbf{end for}
\STATE  $\quad$  Compute $\z_i^b=\sum_{t=(b-1)L+1}^{bL}\g_i^t$
\STATE $\quad$ \textbf{if} $b\geq2$ \textbf{do}
\STATE $\quad$  $\quad$   $\mathbf{v}_i^{b-1}$ = FCC$(\e_i^{b-1} + \z_i^{b-1},\C(\cdot), L, \text{server} )$ 
     \LineComment{\textcolor{blue}{Transmission distributed over $L$ rounds}}
\STATE  $\quad$ $\quad$ Compute $\e_i^{b} = \e_i^{b-1} + \z_i^{b-1} - \mathbf{v}_i^{b-1}$
\STATE \textbf{for server and if }$b\geq 3$ \textbf{do}
    \STATE $\quad$ Compute $\mathbf{v}^{b-2} = \frac{1}{n} \sum_{i=1}^n \mathbf{v}_i^{b-2}$ \LineComment{\textcolor{blue}{Only receive gradients of block $b-2$}}
    \STATE  $\quad$ $\s^{b-2}=$FCC($\mathbf{v}^{b-2}+\hat{\e}^{b-2}$, $\C(\cdot)$, $L$, learner)
    \LineComment{\textcolor{blue}{Transmission distributed over $L$ rounds}}
    \STATE $\quad$ Compute $\hat{\e}^{b-1} = \hat{\e}^{b-2} + \mathbf{v}^{b-2} - \s^{b-2}$
\STATE \textbf{for each learner} $i = 1, \ldots, n$  \textbf{in parallel and if} $b\geq3$ \textbf{do}
\STATE  $\quad$ Update the decision $\w^{b+1}$ according to \eqref{eq:ftfcl-convex} or  \eqref{eq:ftfcl-sconvex} 
    \ENDFOR
\end{algorithmic}
\end{algorithm}
 
 To resolve this dilemma,  we utilize the online compression strategy \citep{Yang2026EfficientDOCO}. By dividing the total $T$ rounds into blocks of size $L$ and updating the decision only at the end of each block, we can effectively distribute the communication cost of each update over the entire block.  At round $t\in[(b-1)L+1,bL]$ within the $b$-th block, each learner $i$ plays the fixed decision $\w^b$ and receives gradient $\g_i^t$. Simultaneously, it transmits the compressed information to the server by performing FCC. Then it updates the local error $\e_i^b$. The server collects the aggregated information at the end of block $b$. In the subsequent block, server applies the EF and employs online compression strategy to disseminate the compressed information back to the learners. 
 
 However, the block-wise transmission mechanism inherently introduces new challenges: since transmitting information between learners and the server requires $L$ rounds, the communication inevitably becomes  \emph{asynchronous} due to transmission latency. In particular, after learners send compressed information to the server through $L$ rounds, it takes another $L$ rounds for transmitting back the aggregated message. As a result, learner $i$ only receives the compressed information $\s^{b-2}$ of block $b-2$ at the end of block $b$.  Reflecting the information lag, the learners update the decision using only the available information up to block $b-2$:
 \begin{equation}\label{eq:ftfcl-convex}
    \w^{b+1}=\arg\min_{\w\in\W}\left\langle\sum_{k=1}^{b-2}\s^{k},\w\right\rangle+\frac{1}{\eta}\Norm{\w}^2.
\end{equation}
If the loss is $\mu$-strongly convex, learners update as follows: 
  \begin{equation}\label{eq:ftfcl-sconvex}
    \w^{b+1}=\arg\min_{\w\in\W}\left\langle\sum_{k=1}^{b-2}\s^{k},\w\right\rangle+\frac{\mu L}{2}\sum_{k=1}^b\Norm{\w-\w^k}^2.
\end{equation}
For the sake of clarity, we present a simplified version of our algorithm, termed Distributed Follow-the-Fast-Compressed-Leader (D-FTFCL), in Algorithm~\ref{alg:2}, with the comprehensive details provided in Appendix~\ref{appendix:detail}. The asynchronous updates pose challenges to the theoretical analysis. To address this, we explicitly characterize the impact of the delayed updates.
 In the following, we establish the guarantees for D-FTFCL.

\begin{theorem}\label{thm:3}
Under the Assumptions~\ref{convex}, \ref{gradient} and \ref{domain}, by setting the learning rate $\eta=\frac{D}{G\sqrt{ L T}}$ and the block size $L = \lceil \frac{1}{\delta} \rceil$, Algorithm~\ref{alg:2} can ensure 
\begin{equation*}
    \EC{R_T}\leq O(\delta^{-1/2}DG\sqrt{T}).
\end{equation*}
\end{theorem}

\begin{theorem}\label{thm:4}
Under the Assumptions~\ref{sconvex}, \ref{gradient} and \ref{domain}, by setting the block size $L = \lceil \frac{1}{\delta} \rceil$, Algorithm~\ref{alg:2} can ensure 
\begin{equation*}
    \EC{R_T}\leq O\left(\mu^{-1}\delta^{-1} (G+\mu D)^2 \log T\right).
\end{equation*}
\end{theorem}
\begin{remark}
Our improved algorithm enjoys $O(\delta^{-1/2}\sqrt{T})$ and $O\left(\delta^{-1}  \log T\right)$ regret bounds for convex and strongly convex loss functions, which match the lower bounds in Theorems~\ref{thm:7} and \ref{thm:8}.
\end{remark}

\begin{remark}
Our theoretical guarantees require setting the block size to $L = \lceil 1/\delta \rceil$ to adequately suppress the compression error. It is worth noting that this choice is derived from the worst-case analysis. In practice, we can employ a smaller $L$ in real-world implementations.
\end{remark}


\section{Application to Distributed Stochastic Convex Optimization}

In this section, we extend our online algorithm to distributed stochastic convex non-smooth optimization with domain constraints. Formally, we consider the following offline distributed optimization problem \citep{nedic2009distributed}
\begin{equation}
 \min_{\x\in\W} f(\x)=\frac{1}{n}\sum_{i=1}^nf_i(\x),   
\end{equation}
where $f_i(\cdot):\W\to\R$ is the loss function of the learner $i$. Here, learners are assumed to have access solely to stochastic estimates of the function value $f_i(\cdot;\xi_i)$ and the local subgradient with sample $\xi_i$ from distribution $\mathcal{D}_i$.  Following the previous work \citep{cutkosky2019anytime}, we introduce the standard assumption outlined below.

\begin{ass}[stochastic subgradient] \label{sgradient} The stochastic subgradient $\g_i^t$ of function $f_{i}(\cdot)$ with random sample $\xi_i^t$ at point $\x^t$ satisfies $\E[\g_i^t]\in \partial f_{i}(\x^t)$ and $\Norm{\g_i^t}^2\leq G^2$.
\end{ass}

As discussed in Section~\ref{doo}, existing studies on distributed optimization typically assume either smooth loss functions or an unconstrained optimization domain. There still remains a critical gap: no work has addressed distributed non-smooth optimization with compressed communication and domain constraints, despite its extensive applications in machine learning tasks \citep{Agarwal2006Portfolio,hazan2016introduction}.  While a recent work by \citet{islamov2025safeef} investigates convex non-smooth optimization, it is limited to the specific setting of safe constraints. Crucially, their methods cannot handle standard domain constraints. Hence, a fundamental question arises: \emph{is it possible to design an algorithm for distributed optimization with compressed communication and domain constraints?}

We provide an affirmative answer for this question by extending our online methods to offline stochastic optimization. An effective approach to solving the stochastic convex  optimization problem is to leverage online learning algorithms  through the anytime online-to-batch (O2B) conversion \citep{cesa2004generalization,shalev2012online,cutkosky2019anytime}, which constructs a sequence of surrogate loss functions using stochastic gradients and applies online learning algorithms to generate decisions. By minimizing the \emph{regret}, we can achieve the optimal convergence rates for stochastic convex optimization problem.

In this work, we propose a distributed variant of the anytime O2B conversion \citep{cutkosky2019anytime}, as summarized in Algorithm~\ref{alg:do2b}\footnote{Given sequences of  weights $\{\alpha_t\}_t$, we use $\alpha_{1:t} = \sum_{k=1}^t\alpha_k$ to represent their summation.}.  To mitigate the compression error in bidirectional compression, we again adopt the FCC technique \citep{huang2022lower}. A key distinction from the online setting is that local learners can communicate with the central server multiple times per update in the offline setting, it eliminates the need for the blocking update technique. Consequently, in the anytime O2B conversion, we can employ our D-FTCL (Algorithm~\ref{alg:1}) as the online method $\A$ and  directly replace the standard compressor $\C(\cdot)$ with the FCC mechanism (Algorithm~\ref{rpc}). At each round $t$, each learner computes a weighted average decision $\x^t=\frac{\sum_{k=1}^t \alpha_k \w^k}{\alpha_{1:t}}$ and queries the local subgradient $\g_i^t$ at $\x^t$. Then, it constructs the surrogate loss according to:
\begin{equation}\label{convex-loss}
    \hat{\ell}^t_i(\w)=\langle \alpha_t \g_i^t, \w \rangle,
\end{equation}
 and feeds it to D-FTCL as the $t$-th loss function. Then, the online method updates the decision according to \eqref{eq:ftcl-convex}. Let $T$ denote the total number of communication rounds. Since the FCC mechanism requires $L$ rounds per update, each learner only performs $K = T/L$ updates. We now establish the \emph{last-iterate} convergence rates for Algorithm~\ref{alg:do2b}. 

\begin{algorithm}[t]
\caption{Distributed O2B Conversion with Compressed Communication}
\label{alg:do2b}
\begin{algorithmic}[1]
\REQUIRE Online method $\A$, weights $\alpha_1, \ldots, \alpha_{T/L}$, compression round $L$, communication round $T$
\STATE Initialize $\w^1=\textbf{0}$, $\A$ as Algorithm~\ref{alg:1}, compressor as Algorithm~\ref{rpc}
\FOR{$t = 1$ to $T/L$}
\FOR{\textbf{each learner} $i=1,...,n$ \textbf{in parallel}}
    \STATE Compute the decision $\x^t = \frac{\sum_{k=1}^t \alpha_k \w^k}{\alpha_{1:t}}$
    \STATE Query subgradient $\g^t_i$ at $\x^t$
    \STATE Construct the surrogate loss $\hat{\ell}^t_i(\w)$ according to \eqref{convex-loss} or \eqref{sconvex-loss}
    \STATE Send  $\hat{\ell}^t_i(\w)$ to $\A$ as $t$-th loss
    \STATE Receive the updated decision $\w^{t+1}$ from $\mathcal{A}$
\ENDFOR
\ENDFOR
\end{algorithmic}
\end{algorithm}
\begin{theorem}\label{thm:5}
    Under the Assumptions~\ref{convex}, \ref{domain} and \ref{sgradient}, configured with the online method $\A$ as Algorithm~\ref{alg:1}, the compression method as Algorithm~\ref{rpc},  $\alpha_t=1$, $\hat{\ell}^t_i(\w)$ according to \eqref{convex-loss}, $\w^{t+1}$ according to \eqref{eq:ftcl-convex}, $\eta = \frac{D}{G\sqrt{K}}$  and compression round $L = \lceil1/\delta\rceil$, for all $\x \in \W$,   Algorithm~\ref{alg:do2b} can ensure 
    \begin{equation*}
        \EC{f(\x^K)-f(\x)} \leq O\left(\frac{DG}{\sqrt{\delta T}}\right),
    \end{equation*}
        where $T=KL$ is  the total communication round.
\end{theorem}
It is not hard to verify that applying the uniform weighting scheme ($\alpha_t = 1$) to strongly convex functions only yields a convergence rate of $O(\delta^{-1} T^{-1}\log T)$, which is suboptimal with respect to $T$. Benefiting from the flexibility of the anytime O2B conversion, we can employ time-varying weights with $\alpha_t=t$ and construct the surrogate loss as:
 \begin{equation}\label{sconvex-loss}
     \hat{\ell}^t_i(\w) = \langle \alpha_t\g_i^t, \w \rangle+\frac{\mu \alpha_t}{2}\Norm{\w-\x^t}^2.
 \end{equation}
 Online method receives the subgradient $\alpha_t\g_i^t$ and updates the decision according to:
 \begin{equation}\label{sconvex-loss-update}
\w^{t+1}=\arg\min_{\w\in\W}\left\langle\sum_{k=1}^t\s^{k},\w\right\rangle+\frac{\mu}{2}\sum_{k=1}^t \alpha_k \Norm{\w-\x^k}^2.
 \end{equation}

\begin{theorem}\label{thm:6}
    Under the Assumptions~\ref{sconvex}, \ref{domain} and \ref{sgradient}, configured with the online method $\A$ as Algorithm~\ref{alg:1}, the compression method as Algorithm~\ref{rpc}, $\alpha_t=t$, $\hat{\ell}^t_i(\w)$ according to \eqref{sconvex-loss}, $\w^{t+1}$ according to \eqref{sconvex-loss-update} and compression round $L = \lceil1/\delta\rceil$, for all $\x \in \W$,  Algorithm~\ref{alg:do2b} can ensure 
    \begin{equation*}
        \EC{ f(\x^K)-f(\x)} \leq O\left(\frac{(\mu D+G)^2}{\mu \delta T}\right),
    \end{equation*}
    where $T=KL$ is  the total communication round.
\end{theorem}

\begin{remark}
    \citet{islamov2025safeef} establish a lower bound of $\Omega(\delta^{-1/2} T^{-1/2})$ for distributed convex non-smooth optimization in the unconstrained setting, where $T$ is the communication round.  Our rate of $O(\delta^{-1/2} T^{-1/2})$ for convex objectives matches this lower bound. Furthermore, in the absence of domain and safety constraints, our method still achieves a more favorable rate than that of \citet{islamov2025safeef}, demonstrating the inherent efficiency of our algorithmic design. For $\mu$-strongly convex functions, our method achieves a convergence rate of $O(\delta^{-1} T^{-1})$, which is optimal with respect to $T$ \cite{Agarwal2012OracleLowerBound}.   Crucially, our results fill a significant gap in the literature by establishing the first convergence guarantees for distributed convex non-smooth optimization with compressed communication under domain constraints.
\end{remark}

\begin{remark}
    In recent years, there have been interests of investigating the sign-based methods \cite{seide20141,koloskova2019decentralized,jiangefficient}, which are advantageous in distributed settings for low communication overhead. If we adopt the (scaled) sign compressor ($\delta = 1/d$) \cite{karimireddy2019error}, our method achieves rates of $O(d^{1/2} T^{-1/2})$ and $O(d T^{-1})$ for convex and strongly convex functions, respectively.
\end{remark}

\section{Conclusion and Future Work}
In this paper, we study distributed online convex optimization (D-OCO) with compressed communication. Firstly,  we propose a communication-efficient method that achieves the regret bounds of $O(\delta^{-1/2}\sqrt{T})$ and $O(\delta^{-1}\log T)$ for convex and strongly convex loss functions, respectively. Secondly, to certify the optimality of our algorithm, we establish lower bounds of $\Omega(\delta^{-1/2}\sqrt{T})$ and $\Omega(\delta^{-1}\log T)$ for convex and strongly convex loss functions, respectively. Finally, we extend our online methods to distributed offline stochastic optimization with domain constraints. We obtain the \emph{first} convergence rates for both convex and strongly convex non-smooth objectives by developing a distributed anytime online-to-batch conversion.  

A promising direction for future work is to investigate whether our method can achieve tighter results for smooth loss functions. It is well-known that optimistic online learning methods \citep{Rakhlin2013Predictable,mohri2016accelerating} can leverage prior knowledge to construct optimistic terms and exploit smoothness to obtain better bounds. However, achieving such improvements in D-OCO with compressed communication presents significant technical challenges. Since each learner receives only compressed information at each round, the compressed gradients alone are insufficient to construct effective optimistic terms.


\bibliography{./ref}
\bibliographystyle{plainnat}

\newpage
\appendix



\section{Additional Discussions on Related Work }\label{decentralizedoco}
\subsection{Compressor}
Generally, compressors can be categorized into two  classes: unbiased compressors \citep{jiang2018linear,tang2018communication,zhang2017zipml} and contractive compressors \citep{seide20141,stich2018sparsified,richtarik20223pc,beznosikov2023biased}. An unbiased compressor satisfies $\E[\C(\x)]=\x$ and $\EC{\Norm{\C(\x) - \x}^2}\leq \omega \Norm{\x}$ for any input $\x \in \R^d$ and $\omega \geq 0$.  The contractive  compressor satisfies $\EC{\Norm{\C(\x) - \x}^2}\leq (1-\delta)\Norm{\x}$ and $\delta \in (0,1]$.  We provide two examples of compressor.
\begin{itemize}
    \item $\text{Rand}$-$K$. Randomly selecting $k$ out of $d$ coordinates yields compressor with the compression ratio of $\delta = \frac{k}{d}$. 
    \item (Scaled) sign. Let $\text{Sign}(\x)$ denote the sign of a vector $\x \in \R^d$. Outputting $\C(\x) = \frac{\Norm{\x}_1}{d}\text{Sign}(\x)$ leads to a compression ratio of $\delta = \frac{1}{d}$. 
\end{itemize}

\section{Proof of Theorems}\label{appendix:proof}

\subsection{Proof of Theorem \ref{thm:1}}
First, we introduce several notations. We denote $f_t(\w) = \frac{1}{n}\sum_{i=1}^n f_i^t(\w)$,  $\mathbf{v}^t=\frac{1}{n}\sum_{i=1}^n\mathbf{v}_i^t$ and $\mathbf{g}^t=\frac{1}{n}\sum_{i=1}^n\mathbf{g}_i^t$.  For $\hat{\e}^k$  and $\e^t$, by summing up  from $k=1$ to $t$, we have 
\begin{align*}
     \e^t&=\sum_{k=1}^{t-1}\mathbf{v}^k-\g^k, \qquad \hat{\e}^t=\sum_{k=1}^{t-1}\s^{k}-\mathbf{v}^k.
\end{align*}
Then,  we define several virtual decisions
\begin{equation*}
    \hat{\w}^{t+1}=\arg\min_{\w\in\W}\left\langle\sum_{k=1}^t\g^k,\w\right\rangle+\frac{1}{\eta}\Norm{\w}^2, \quad \y^{t+1}=\arg\min_{\y\in\W}\left\langle\sum_{k=1}^t\mathbf{v}^k,\y\right\rangle+\frac{1}{\eta}\Norm{\y}^2.
\end{equation*}

Note that $\hat{\w}^{t+1}$ represents the decision updated with the exact gradient, whereas $\y^{t+1}$ represents the decision updated with information that learners transmit to the server. Then we introduce some useful lemmas.
\begin{lem}[Lemma 6.6 in \citet{Garber2016}] \label{lem:gh16}
Let $\{\ell_t(\w)\}_{t=1}^T$ be a sequence of functions and 
$\w_t^\star \in \arg\min_{\w \in \W} \sum_{k=1}^t \ell_k(\w)$ for any $t \in [T]$. 
Then, it holds that
\begin{equation*}
    \sum_{t=1}^T \ell_t(\w_t^\star) - \min_{\w \in \W} \sum_{t=1}^T \ell_t(\w) \le 0.
\end{equation*}
\end{lem}

\begin{lem}[Lemma 5 in \citet{Duchi2011}] \label{lem:duchi11}
Let $\Pi_{\W}(\mathbf{u}, \eta) = \arg\min_{\w \in \W} \langle \mathbf{u}, \w \rangle + \frac{1}{\eta} \|\w\|^2$.
For any $\mathbf{u}, \mathbf{v} \in \R^d$, we have
\begin{equation*}
\|\Pi_{\W}(\mathbf{u}, \eta) - \Pi_{\W}(\mathbf{v}, \eta)\| \le \frac{\eta}{2} \|\mathbf{u} - \mathbf{v}\|.    
\end{equation*}
\end{lem}
Now, we are ready to derive the regret bound.
\begin{align*}
    \EC{R_T}&=\EC{\frac{1}{n}\sum_{t=1}^T\sum_{i=1}^nf_i^t(\w^t)-\frac{1}{n}\sum_{t=1}^T\sum_{i=1}^nf_i^t(\w)}\\
    &\leq\frac{1}{n}\sum_{t=1}^T\sum_{i=1}^n\E_{\C}\left[\langle\g_i^t,\w^t-\w\rangle\right]\\
    &=\sum_{t=1}^T\E_{\C}\left[\langle\g^t,\w^t-\w\rangle\right]\\
    &\leq \sum_{t=1}^T\EC{\langle\g^t,\w^t-\hat{\w}^{t}+\hat{\w}^{t}-\w\rangle}\\
    &=\sum_{t=1}^T\EC{\langle \g^t,\hat{\w}^{t}-\w \rangle +\langle\g^t,\w^t-\hat{\w}^{t}\rangle}\\
    &\leq \underbrace{\sum_{t=1}^T\EC{\langle\g^t,\hat{\w}^{t}-\w \rangle}}_{\mathtt{TERM}\text{-}\mathtt{A}} +  \underbrace{\sum_{t=1}^TG\EC{\Norm{\w^t-\hat{\w}^{t}}}}_{\mathtt{TERM}\text{-}\mathtt{B}},
\end{align*}
where the last inequality is due to $\langle \x,\y\rangle \leq \Norm{\x}\Norm{\y}$ and $\Norm{\g^t}=\Norm{\frac{1}{n}\sum_{i=1}^n\g_i^t}\leq G$.
Remarkably, $\mathtt{TERM}\text{-}\mathtt{A}$ is the regret bound of the decisions updated by using the uncompressed gradients and $\mathtt{TERM}\text{-}\mathtt{B}$ is the error of the compression. In the following, we give the bound of this two term, respectively.

As for $\mathtt{TERM}\text{-}\mathtt{A}$, we have 
\begin{align*}
    \sum_{t=1}^T\langle\g^t,\hat{\w}^{t}-\w\rangle=&\sum_{t=1}^T\langle\g^t,\hat{\w}^{t}-\hat{\w}^{t+1}+\hat{\w}^{t+1}-\w\rangle\\
    =&\sum_{t=1}^T\langle\g^t,\hat{\w}^{t}-\hat{\w}^{t+1}\rangle+\sum_{t=1}^T\langle\g^t,\hat{\w}^{t+1}-\w\rangle\\
    \leq &\sum_{t=1}^TG\Norm{\hat{\w}^{t}-\hat{\w}^{t+1}}+\sum_{t=1}^T\langle\g^t,\hat{\w}^{t+1}-\w\rangle.
\end{align*}

We define $\ell_1(\w)=\langle \g^1,\w\rangle+\frac{1}{\eta}\Norm{\w}^2$ and $\ell_t(\w)=\langle\g^t,\w\rangle$. We can verify that  
$\hat{\w}^{t+1} = \arg\min_{\w \in \W} \sum_{k=1}^t \ell_k(\w)$. By using Lemma \ref{lem:gh16}, we can derive 
\begin{equation}\label{eq:0}
\begin{aligned}
    \sum_{t=1}^T\langle\g^t,\hat{\w}^{t+1}-\w\rangle+\frac{\Norm{\hat{\w}^2}^2-\Norm{\w}^2}{\eta}\leq&\sum_{t=1}^T\ell_t(\hat{\w}^{t+1})-\ell_t(\w)\leq 0.
\end{aligned}    
\end{equation}

Therefore, we can obtain
\begin{equation}\label{eq:16}
\begin{aligned}
    \sum_{t=1}^T\langle\g^t,\hat{\w}^{t}-\w\rangle=&\sum_{t=1}^T\langle\g^t,\hat{\w}^{t}-\hat{\w}^{t+1}\rangle+\sum_{t=1}^T\langle\g^t,\hat{\w}^{t+1}-\w\rangle\\
    \leq&\sum_{t=1}^T\Norm{\g^t}\Norm{\hat{\w}^t-\hat{\w}^{t+1}}+\frac{\Norm{\hat{\w}^2}^2-\Norm{\w}^2}{\eta}\\
    \leq&\frac{\eta G^2 T}{2}+\frac{D^2}{\eta},
\end{aligned}
\end{equation}
where the first inequality is due to (\ref{eq:0}).

In the following, we give the bound of $\mathtt{TERM}\text{-}\mathtt{B}$.
\begin{align*}
    \EC{\Norm{\w^t-\hat{\w}^{t}}}&=\EC{\Norm{\w^t-\y^{t}+\y^{t}-\hat{\w}^t}}\\
    &\leq \EC{\Norm{\w^t-\y^{t}}}+\EC{\Norm{\y^{t}-\hat{\w}^t}}\\
    &\leq \frac{\eta}{2}\left(\EC{\Norm{\sum_{k=1}^{t-1}\s^{k}-\mathbf{v}^k}}+\EC{\Norm{\sum_{k=1}^{t-1}\mathbf{v}^k-\g^k}}\right)\\    
    &=\frac{\eta}{2}\left(\EC{\Norm{\e^t}}+\EC{\Norm{\hat{\e}^t}}\right),
\end{align*}
where the second inequality is due to Lemma \ref{lem:duchi11}.
To give the bound of errors, we introduce the following lemma.
\begin{lem}\label{lem:1}
    Under the Assumptions \ref{gradient} and \ref{domain},  at any round $t$ of D-FTCL, the norm of the errors is bounded by
    \begin{align*}
       \EC{\Norm{\e^{t+1}}^2}&\leq\frac{4(1-\delta)G^2}{\delta^2}, \quad \EC{\Norm{\hat{\e}^{t+1}}^2}\leq\frac{160(1-\delta)G^2}{\delta^4}.\\
    \end{align*}
\end{lem}

By using Lemma \ref{lem:1}, we have 
\begin{align*}
    \EC{\Norm{\w^t-\hat{\w}^{t}}}&\leq \frac{\eta}{2}\left(\EC{\Norm{\hat{\e}^{t-1}}}+\EC{\Norm{\e^{t-1}}}\right)\\
    &\leq \frac{\eta}{2}\left(\frac{13\sqrt{1-\delta}G}{\delta^2}+\frac{2\sqrt{1-\delta}G}{\delta}\right) \leq \frac{15\eta G}{2\delta^2}.
\end{align*}

By setting the learning rate $\eta = \frac{\delta D}{G\sqrt{T}}$, the final regret bound is 
\begin{align*}
    \EC{R_T} &\leq \sum_{t=1}^T\EC{\langle\g^t,\hat{\w}^{t}-\w\rangle} +\EC{G\Norm{\w^t-\hat{\w}^{t}}}\\
    &\leq \frac{\eta G^2 T}{2}+\frac{D^2}{\eta}+\frac{15\eta G^2}{2\delta^2}\\
    &\leq \frac{DG \delta \sqrt{T}}{2}+\frac{DG\sqrt{T}}{\delta }+\frac{15DG\sqrt{T}}{2\delta}\\
    &\leq O(\delta^{-1}DG\sqrt{T}).
\end{align*}

\subsection{Proof of Theorem \ref{thm:2}}
Similar to the previous proof, we define two decisions.

\begin{align*}
    \y^{t+1}&=\arg\min_{\y\in\W}\left\langle\sum_{k=1}^t\mathbf{v}^k,\y\right\rangle+\frac{\mu}{2}\sum_{k=1}^{t}\Norm{\y-\w^k}^2\\
    &=\arg\min_{\y\in\W}\left\langle\sum_{k=1}^t\mathbf{v}^k - \mu \w^k,\y\right\rangle+\frac{\mu t}{2}\Norm{\y}^2.
\end{align*}

\begin{align*}
    \hat{\w}^{t+1}&=\arg\min_{\w\in\W}\left\langle\sum_{k=1}^t\g^k,\w\right\rangle+\frac{\mu}{2}\sum_{k=1}^{t}\Norm{\w-\w^k}^2\\
    &=\arg\min_{\w\in\W}\left\langle\sum_{k=1}^t\g^k - \mu \w^k,\w\right\rangle+\frac{\mu t}{2}\Norm{\w}^2.
\end{align*}
We begin to derive the regret bound, where the key difference is to utilize the strong convexity to improve the dependence on $T$.
\begin{align*}
    \EC{R_T}&=\EC{\frac{1}{n}\sum_{t=1}^T\sum_{i=1}^n f_i^t(\w^t)-f_i^t(\w)}\\
    &\leq\frac{1}{n}\sum_{t=1}^T\sum_{i=1}^n\E_{\C}\left[\langle\g_i^t,\w^t-\w\rangle-\frac{\mu}{2}\Norm{\w^{t}-\w}\right]\\
    &=\sum_{t=1}^T\E_{\C}\left[\langle\g^t,\w^t-\w\rangle-\frac{\mu}{2}\Norm{\w^{t}-\w}\right]\\
    &=\sum_{t=1}^T\E_{\C}\left[\langle\g^t,\w^t-\hat{\w}^{t}+\hat{\w}^{t}-\w\rangle-\frac{\mu }{2}\Norm{\w^{t}-\w}\right]\\
    &=\sum_{t=1}^T\E_{\C}\left[\langle\g^t,\w^t-\hat{\w}^{t}\rangle\right]+\E\left[\langle\g^t,\hat{\w}^{t}-\w\rangle-\frac{\mu }{2}\Norm{\w^{t}-\w}\right]\\
    &\leq \underbrace{\sum_{t=1}^T\E\left[\langle\g^t,\hat{\w}^{t}-\w\rangle-\frac{\mu }{2}\Norm{\w^{t}-\w}\right]}_{\mathtt{TERM}\text{-}\mathtt{A}} + \underbrace{\sum_{t=1}^TG\Norm{\w^t-\hat{\w}^{t}}}_{\mathtt{TERM}\text{-}\mathtt{B}},
\end{align*}
where the first inequality is due to Assumption \ref{sconvex} and the last inequality is due to $\langle \x,\y\rangle\leq \Norm{\x}\Norm{\y}.$

We first give the bound of $\mathtt{TERM}\text{-}\mathtt{A}$. We define $\ell_t(\w)=\langle \g^t,\w\rangle+\frac{\mu }{2}\Norm{\w-\w^t}^2$ and $\hat{\w}^{t+1}=\arg\min_{\w\in\W}\sum_{k=1}^t \ell_k(\w)$ in the strongly convex case. Then, it is not hard to verify that
\begin{equation}\label{eq:1}
\begin{aligned}
    &\sum_{t=1}^{T}\langle\g^t,\hat{\w}^{t}-\w\rangle-\frac{\mu }{2}\Norm{\w^{t}-\w}\\
    =&{}\sum_{t=1}^{T}\langle\g^t,\hat{\w}^{t}-\hat{\w}^{t+1}+\hat{\w}^{t+1}-\w\rangle-\frac{\mu}{2}\Norm{\w^{t}-\w}\\
    \leq& \sum_{t=1}^{T}\ell_t(\hat{\w}^{t+1})-\ell_t(\w)+G\Norm{\hat{\w}^{t}-\hat{\w}^{t+1}}\\
    \leq& \sum_{t=1}^{T} G\Norm{\hat{\w}^{t}-\hat{\w}^{t+1}}.
\end{aligned}
\end{equation}

Next, we bound the term $G \Norm{\hat{\w}^{t}-\hat{\w}^{t+1}}$. We define $F_t(\w)=\sum_{k=1}^t\ell_k(\w)$, and  $F_t(\w)$ is a $(t\mu)$-strongly convex function and $\hat{\w}^{t+1}=\arg\min_{\w\in\W}F_t(\w)$. We introduce the following lemma from \citet{hazan2012projection}.

\begin{lem}\label{lem:3} 
For any $\mu$-strongly convex function $f(\x): \W \to \R$ and any $\x \in \W$, it holds that
\begin{equation}
\frac{\mu}{2}\|\x - \x^\star\|^2 \le f(\x) - f(\x^\star),
\end{equation}
where $\x^\star = \arg\min_{\x \in \W} f(\x)$.
\end{lem}

For any $\x,\y\in\W$, we have
\begin{align*}
    |\ell_t(\x)-\ell_t(\y)|&\leq\langle\nabla\ell_t(\x),\x-\y\rangle\\
    &\leq \Norm{\nabla \ell_t(\x)}\Norm{\x-\y}\\
    &\leq (G+\mu D)\Norm{\x-\y}.
\end{align*}

By using Lemma \ref{lem:3}, we can derive
\begin{align*}
    G\Norm{\hat{\w}^{t+1}-\hat{\w}^t}^2\leq&\frac{2G}{t\mu}(F_t(\hat{\w}^t)-F_t(\hat{\w}^{t+1}))\\
    \leq&\frac{2G}{t\mu}(F_{t-1}(\hat{\w}^t)-F_{t-1}(\hat{\w}^{t+1})+\ell_t(\hat{\w}^t)-\ell_t(\hat{\w}^{t+1}))\\
    \leq &\frac{2G}{t\mu}(\ell_t(\hat{\w}^t)-\ell_t(\hat{\w}^{t+1}))\\
    \leq &\frac{2G}{t\mu}(G+\mu D)\Norm{\hat{\w}^{t+1}-\hat{\w}^t},
\end{align*}
which means
\begin{equation}\label{eq:2}
    G\Norm{\hat{\w}^{t+1}-\hat{\w}^t}\leq \frac{2G}{t\mu}(G+\mu D).
\end{equation}

By using Lemma \ref{lem:duchi11} and Lemma \ref{lem:1}, we have 
\begin{align*}
    \EC{\Norm{\w^t-\hat{\w}^{t}}}&\leq \frac{1}{t\mu}\left(\EC{\Norm{\hat{\e}^{t-1}}}+\EC{\Norm{\e^{t-1}}}\right)\\
    &\leq \frac{1}{t\mu}\left(\frac{13\sqrt{1-\delta}G}{\delta^2}+\frac{2\sqrt{1-\delta}G}{\delta}\right)\leq \frac{15 G}{t\mu\delta^2}.
\end{align*}

Finally, we can derive the regret bound
\begin{align*}
    \EC{R_T}&\leq \sum_{t=1}^{T}\frac{2G}{t\mu}(G+\mu D)+ \frac{15 G}{t\mu\delta^2}\leq \frac{15(G+\mu D)^2}{\mu \delta^2}\log T=O(\delta^{-2}\log T).
\end{align*}


\subsection{Proof of Theorem \ref{thm:7}}

In this section, we present the proof of the lower bound for D-OCO with compressed communication. We follow \citet{huang2022lower} and assume two protocols: (i) all learners can only communication with the central server and cannot exchange information with one another; (ii) all updates are synchronized,  which means  all learners update their decisions using the same information received from the central server, ensuring that the decisions across all learners are consistent.

To characterize the effect of the compressor, we employ the idea of maximizing communication delay by using a specific compressor \citep{huang2022lower}. In our construction, we consider the Random Gossip compressor $\C(\cdot)$ \citep{koloskova2019decentralized}, which outputs $Q(\x)=\x$ with probability $\delta \in(0,1]$ and $Q(\x)=\textbf{0}$ otherwise. Under this scheme, the expected number of communication rounds required for a learner to successfully transmit information to the central server is $\lceil1/\delta\rceil$. We assume the server does not use the compressor, which is unidirectional compression. As pointed out by \citet{huang2022lower}, the lower bound for algorithms that admit bidirectional compression is greater than or equal to that with unidirectional compression. Therefore, we do not employ the compressor on the server.  As noted by \citet{huang2022lower}, lower bounds for bidirectional compression are at least as large as those for unidirectional compression. To simplify the analysis, we consider unidirectional compression where only learners compress their communications, while the server transmits without compression. 

Let $K=\lceil 1/\delta\rceil, Z=\lfloor (T-1)/K \rfloor, c_0=0$ and $c_{Z+1}=T$. The total $T$ rounds can be divided into the following $Z+1$ intervals $[c_{0}+1,\,c_{1}],\,[c_{1}+1,\,c_{2}],\,\ldots,\,[c_{Z}+1,\,c_{Z+1}]$. To maximize the impact of the communication on the regret, we set the losses of each learner $i$ is same in each interval.  

Specifically, we assume $m=\lfloor n/2 \rfloor$ and set the loss functions as
\begin{align*}
    f_{1}^t(\mathbf{w}) = \cdots = f_{m}^t(\mathbf{w}) = 0,
\end{align*}
 for $t \in [c_{i}+1,\,c_{i+1}], i\in\{0,...,Z\}$, while the other loss functions are set $f^{t}_{m+1}(\mathbf{w}) = \cdots = f_{n}^{t}(\mathbf{w}) = h_i(\w)$. In this way, the global loss function is
 \begin{equation*}
     f_t(\w)=\frac{n-m}{n}h_i(\w).
 \end{equation*}
 We  independently select $h_i(\w)=\langle\z_i,\w\rangle$ for any $i\in\{0,...,Z\}$, where the
coordinates of $\z_i$ are $\pm G/\sqrt{d}$ with probability $0.5$ and the feasible domain $\mathcal{X}=[-D/2\sqrt{d},D/2\sqrt{d}]^d$.
 According to the above discussion, the
decisions $\x^t$ for any $t \in [c_{i}+1,\,c_{i+1}], i\in\{0,...,Z\}$ are made before the function $h_i(\w)$ can be revealed to the local learner. Then we can derive 
\begin{equation}\label{eq:3}
    \begin{aligned}
    \EC{R_T}=&\EC{\frac{1}{n}\sum_{t=1}^T\sum_{i=1}^n f^t_i(\w^t)-\min_{\w\in\W}\frac{1}{n}\sum_{t=1}^T\sum_{i=1}^nf^t_i(\w)}\\
    =&\frac{n-m}{n}\EC{
\sum_{i=0}^{Z}\sum_{t=c_i+1}^{c_{i+1}} h_i(\mathbf{w}^t)
- \min_{\w\in\W}\sum_{i=0}^{Z}\sum_{t=c_i+1}^{c_{i+1}} h_i(\mathbf{w})}\\
    =&\frac{n-m}{n}\EC{
\sum_{i=0}^{Z}\sum_{t=c_i+1}^{c_{i+1}} \langle\z_i,\mathbf{w}^t\rangle
- \min_{\w\in\W}\sum_{i=0}^{Z}\sum_{t=c_i+1}^{c_{i+1}} \langle\z_i,\mathbf{w}\rangle}\\
    =&-\frac{n-m}{n}\EC{\min_{\w\in\W}\sum_{i=0}^{Z}\sum_{t=c_i+1}^{c_{i+1}} \langle\z_i,\mathbf{w}\rangle}\\
                =&-\frac{n-m}{n}\EC{\min_{\w\in\W} \langle\sum_{i=0}^{Z}(c_{i+1}-c_i)\z_i,\mathbf{w}\rangle},
\end{aligned}
\end{equation}
where the fourth equality is due to  $\mathbb{E}[\langle \mathbf{z}_i,\mathbf{w}^t\rangle]=0$ for $\forall t\in[c_i+1,c_{i+1}]$.

Then, we denote $\epsilon_{01},...,\epsilon_{0d},...,\epsilon_{Z1},...,\epsilon_{Zd}$ be the coordinates of $\z_1,...,\z_Z$, which are identically distributed variables with $\mathbb{P}(\epsilon_{ij}=\pm1)=1/2$ for $i\in\{0,...,Z\}$ and $j\in\{1,...,d\}$.
By using the Khintchine inequality on (\ref{eq:3}), we have
\begin{equation}\label{eq:4}
    \begin{aligned}
   \EC{R_T}=&-\frac{n-m}{n} \mathbb{E}_{\epsilon_{01},\ldots,\epsilon_{Zd}}\left[\sum_{j=1}^d -\frac{D}{2\sqrt{d}}\left|\sum_{i=1}^Z(c_{i+1}-c_i)\frac{\epsilon_{ij}G}{\sqrt{d}}\right|\right]\\
    =&\frac{n-m}{n}\frac{DG}{2} \mathbb{E}_{\epsilon_{01},\ldots,\epsilon_{Zd}}\left[\left|\sum_{i=0}^Z(c_{i+1}-c_i)\epsilon_{i1}\right|\right]\\
    \geq&\frac{n-m}{n}\frac{DG}{2\sqrt{2}} \sqrt{\sum_{i=0}^{Z}(c_{i+1}-c_{i})^2}\\
\geq&\frac{n-m}{n}\frac{DG}{2\sqrt{2}} \sqrt{\frac{(c_{Z+1}-c_{0})^2}{Z+1}}\\
\geq&\frac{DGT}{4\sqrt{2Z+2}}\\
\geq &\frac{DGT}{4\sqrt{2\delta T+2-2\delta}},
\end{aligned}
\end{equation}
where the second inequality is due to the Cauchy-Schwarz inequality.

If $2-2\delta \leq 2\delta T$, we can obtain
\begin{equation}
    \begin{aligned}
   \EC{R_T}
\geq &\frac{DGT}{4\sqrt{2\delta T+2-2\delta}}
\geq \frac{DG\sqrt{T}}{8\sqrt{\delta}}.
\end{aligned}
\end{equation}

\subsection{Proof of Theorem \ref{thm:8}}

This proof is similar to the proof of Theorem \ref{thm:8}. The key modification is to construct new loss functions to utilize the strong convexity and we choose the feasible domain $\W=[0,D/\sqrt{d}]^d$

Specifically, we set the loss functions as
\begin{align*}
    f^{t}_1(\mathbf{w}) = \cdots = f^{t}_m(\mathbf{w}) = \frac{\mu}{2}\Norm{\w}^2,
\end{align*}
 for $t \in [c_{i}+1,\,c_{i+1}], i\in\{0,...,Z\}$, while the other loss functions are set 
 \begin{equation*}
f^{t}_{m+1}(\mathbf{w}) = \cdots = f^{t}_n(\mathbf{w}) = \frac{\mu}{2}\Norm{\w-\frac{D\z_i}{\sqrt{d}}}^2, 
 \end{equation*}
 where $\z_i$ is sampled from the distribution: $\text{Pr}(\z_i=\textbf{1})=p$, $\text{Pr}(\z_i=\textbf{0})=1-p$ and $p\in[0,1]$.
 In this way, the global loss function in round $t\in [c_i+1,c_{i+1}], i\in\{0,...,Z\}$ is
 
\begin{align*}
     f_t(\w)&=\frac{\mu m}{2n}\Norm{\w}^2+\frac{\mu (n-m)}{2n}\Norm{\w-\frac{D\z_i}{\sqrt{d}}}^2\\
            &=\frac{\mu}{2}\Norm{\w}^2+\frac{\mu(n-m)D^2}{2nd}\Norm{\z_i}^2-\frac{\mu(n-m)D}{\sqrt{d}}\langle \w,\z_i\rangle.
\end{align*}

By taking the expectation on $\z_i$, we have
\begin{align*}
     \E_{\z_i}[f_t(\w)]&=\frac{\mu}{2}\Norm{\w}^2+\frac{\mu(n-m)D^2}{2nd}\langle\textbf{1},\textbf{p}\rangle-\frac{\mu(n-m)D}{\sqrt{d}}\langle \w,\textbf{p}\rangle\\
     &=\frac{\mu}{2}\Norm{\mathbf{w} - \frac{(n - m)D\mathbf{p}}{n\sqrt{d}} }+\frac{\mu (n - m) D^2}{2nd} \left\langle \textbf{1} - \frac{n-m}{n}\mathbf{p}, \mathbf{p} \right\rangle,
\end{align*}
where $\p=[p,...,p]^{\top}\in \R^d$. We denote $H(\w)=\E_{\z_i}[f_i(\w)]$ and $\w^\star=\arg\min_{\w\in\W}F(\w)=\frac{(n - m)D\mathbf{p}}{n\sqrt{d}}\in\W$, which implies $H(\w)-H(\w^\star)\geq 0$. Then we derive the lower bound for strongly convex loss functions.

\begin{equation}\label{eq:5}
    \begin{aligned}
    \EC{R_T}=&\EC{\frac{1}{n}\sum_{t=1}^T\sum_{i=1}^n f_t^i(\w^t)-\min_{\w\in\W}\frac{1}{n}\sum_{t=1}^T\sum_{i=1}^nf_t^i(\w)}\\
    =&\EC{
\sum_{i=0}^{Z}\sum_{t=c_i+1}^{c_{i+1}} h_i(\mathbf{w}^t)
- \min_{\w\in\W}\sum_{i=0}^{Z}\sum_{t=c_i+1}^{c_{i+1}} h_i(\mathbf{w})}\\
    =&\EC{
\sum_{i=0}^{Z}\sum_{t=c_i+1}^{c_{i+1}} H(\mathbf{w}^t)
- \min_{\w\in\W}\sum_{i=0}^{Z}\sum_{t=c_i+1}^{c_{i+1}} H(\mathbf{w})}\\
    \geq&\EC{
\sum_{i=0}^{Z}\sum_{t=c_i+1}^{c_{i+1}} H(\mathbf{w}^t)
- \sum_{i=0}^{Z}\sum_{t=c_i+1}^{c_{i+1}} H(\mathbf{w}^\star)},
\end{aligned}
\end{equation}
where the last inequality is due to $\E\left[\sum_{i=0}^{Z}\sum_{t=c_i+1}^{c_{i+1}}H(\w^\star)\right]\geq \E\left[\min_{\w\in\W}\sum_{i=0}^{Z}\sum_{t=c_i+1}^{c_{i+1}} H(\mathbf{w})\right]$.

Now, our goal is to derive a lower bound of the right term of (\ref{eq:5}). We introduce a lemma from the proof of Theorem 4 in \citet{JMLR:2025:Wan}, which establish the lower bound for decentralized online convex optimization.

\begin{lem}[\citet{JMLR:2025:Wan}]\label{lem:wan}
    For an online convex optimization with the domain $\W=[0,D/\sqrt{d}]^d$, the length of interval is $K$ and the interval number $Z=\lfloor (T-1)/K \rfloor$, $M=\lfloor\log_{16}(15Z+16)-1\rfloor\geq 1$, the lower bound of  (\ref{eq:5}) is
    \begin{equation*}
        \sum_{i=0}^{Z}\sum_{t=c_i+1}^{c_{i+1}} H(\mathbf{w}^t)
- \sum_{i=0}^{Z}\sum_{t=c_i+1}^{c_{i+1}} H(\mathbf{w}^\star)\geq \frac{\mu MK(n-m)^2D^2}{16^{4}n^2}.
    \end{equation*}
\end{lem}

By using lemma \ref{lem:wan} and setting $16\delta^{-1}+1\leq T$, we can obtain the following lower bound
\begin{align*}
    \EC{R_T}\geq& \frac{\mu MK(n-m)^2D^2}{16^{4}n^2}\\
    \geq &  \frac{\mu D^2(\log_{16}(15Z+16)-2)}{4\cdot 16^{4} \cdot \delta }\\
    \geq & \frac{\mu D^2(\log_{16}(15\delta(T-1)+16)-2)}{4\cdot 16^{4} \cdot \delta }\\
    \geq & \frac{\mu D^2(\log_{16}(15\delta T)-2)}{4\cdot 16^{4} \cdot \delta }.
\end{align*}

\textbf{Additional discussion. } Note that the upper regret bounds of D-OCO algorithms generally hold for all compressors satisfying Definition \ref{compress}. Therefore, although lower bounds in Theorems \ref{thm:7} and \ref{thm:8} are constructed based on a specific compressor instantiation, they are sufficient to prove the tightness of the upper bounds in general.


\subsection{Proof of Theorem \ref{thm:3}}\label{appendix:detail}

We first provide a more detailed version of D-FTFCL in Algorithm \ref{alg:detail}.

 \begin{algorithm}[t]
\caption{Distributed Follow-the-Fast-Compressed-Leader (D-FTFCL)}
\label{alg:detail}
\begin{algorithmic}[1]
\REQUIRE $\w^1 =\textbf{0}$, $\e_i^1 = \textbf{0}$, $\hat{\e}^1 = \textbf{0}$,  learning rate $\eta$, compressor $\C(\cdot)$, block size $L$
\FOR{$b = 1, \ldots, T/L$}
\STATE \textbf{for each learner} $i = 1, \ldots, n$ \textbf{in parallel}
\STATE  $\quad$ If $b\geq2$, set $\rr_{i,b-1}^k = \textbf{0}, k=1$
\STATE $\quad$ \textbf{for} $t = (b-1)L+1, \ldots, bL$ \textbf{do}
\STATE  $\quad$   $\quad$  Play the decision $\w^b$ and receive gradient $\g_i^t$
\STATE  $\quad$   $\quad$ If $b\geq2$
\STATE  $\quad$   $\quad$ $\quad$ Compute $ \C(\z_i^{b-1} - \rr_{i,b-1}^k)$ and send to the central server
\STATE  $\quad$   $\quad$ $\quad$ Update $\rr_{i,b-1}^{k+1} = \rr_{i,b-1}^{k} +  \C(\z_i^{b-1} - \rr_{i,b-1}^k)$ and set $k=k+1$
\STATE  $\quad$  $\quad$ \textbf{end if}
\STATE  $\quad$  \textbf{end for}
\STATE  $\quad$  Compute $\z_i^b=\sum_{t=(b-1)L+1}^{bL}\g_i^t$ and $\vv_i^{b-1} = \rr_{i,b-1}^{L+1}$ 
\STATE  $\quad$  Compute $\e_i^{b} = \e_i^{b-1} + \z_i^{b-1} - \mathbf{v}_i^{b-1}$
\STATE \textbf{for server and if }$b\geq 3$ \textbf{do}
\STATE $\quad$ Compute $\mathbf{v}^{b-2} = \frac{1}{n} \sum_{i=1}^n \mathbf{v}_i^{b-2}$ \LineComment{\textcolor{blue}{Only receive gradients of block $b-2$ at block $b$}}
\STATE  $\quad$ Set $\rr_{b-2}^k = \textbf{0}, k=1$
\STATE $\quad$ \textbf{for} $t = (b-1)L+1, \ldots, bL$ \textbf{do}
\STATE  $\quad$  $\quad$ Compute $\C(\mathbf{v}^{b-2}+\hat{\e}^{b-2} - \rr_{b-2}^k)$ and send to all local learners
\STATE   $\quad$ $\quad$ Update $\rr_{b-2}^{k+1} = \rr_{b-2}^{k} + \C(\mathbf{v}^{b-2}+\hat{\e}^{b-2} - \rr_{b-2}^k)$ and set $k=k+1$
\STATE  $\quad$ \textbf{end for}
    \STATE  $\quad$   Compute $\s^{b-2}= \rr_{b-2}^{L+1}$ 
    \STATE $\quad$   Update $\hat{\e}^{b-1} = \hat{\e}^{b-2} + \mathbf{v}^{b-2} - \s^{b-2}$
\STATE \textbf{for each learner} $i = 1, \ldots, n$  \textbf{in parallel and if} $b\geq3$ \textbf{do}
\STATE  $\quad$ Receive total information $\s^{b-2}$ and update the decision $\w^{b+1}$ according to \eqref{eq:ftfcl-convex} or  \eqref{eq:ftfcl-sconvex} 
    \ENDFOR
\end{algorithmic}
\end{algorithm}

Due to the asynchronous communication between the learners and the server, each learner experiences a two-block delay during aggregation. Specifically, for the gradient computed in block $b$, it takes one block for the learner to transmit the information to the server, and another block for the server to aggregate and broadcast the updated information back to the learners. Consequently, at the end of block $b$, each learner updates its model using the gradient information from block $b-2$. 

We first define several decisions.
\begin{equation*}
\hat{\w}^{b+1}=\arg\min_{\w\in\W}\left\langle \sum_{k=1}^b\z^k,\w\right\rangle+\frac{1}{\eta}\Norm{\w}^2,
\end{equation*}
where $\z^b = \frac{1}{n}\sum_{i=1}^n\z_i^b$.
It is not hard to verify that $\hat{\w}^{b+1}$ is the decision with the exact gradient when there is no delay in transmission.
\begin{equation*}
\y^{b+1}=\arg\min_{\y\in\W}\left\langle\sum_{k=1}^{b-2}\mathbf{v}^k,\y\right\rangle+\frac{1}{\eta}\Norm{\y}^2,  \quad \tilde{\w}^{b+1}=\arg\min_{\w\in\W}\left\langle\sum_{k=1}^{b-2}\z^k,\w\right\rangle+\frac{1}{\eta}\Norm{\w}^2.
\end{equation*}
We also have
\begin{align*}
  \e^{b+1}&=\sum_{k=1}^{b}\mathbf{v}^k-\z^k, \qquad \hat{\e}^{b+1}=\sum_{k=1}^{b}\s^{k}-\mathbf{v}^k.
\end{align*}
In the following, we begin to derive the regret 
\begin{align*}
    \EC{R_T}&\leq\frac{1}{n}\sum_{t=1}^T\sum_{i=1}^n\EC{\langle\g_i^t,\w^t-\w\rangle}\\
    &=\frac{1}{n}\sum_{b=1}^K\sum_{t=(b-1)L}^{bL}\sum_{i=1}^n\EC{\langle\g_i^t,\w^b-\w\rangle}\\
    &=\sum_{b=1}^K\sum_{t=(b-1)L}^{bL}\EC{\langle\g^t,\w^b-\w\rangle}\\
    &\leq \sum_{b=1}^K\EC{\langle\z^b,\w^b-\hat{\w}^b+\hat{\w}^b-\w\rangle}\\
    &=\sum_{b=1}^K\EC{\langle \z^b,\hat{\w}^{b}-\w\rangle} +\EC{\langle\z^b,\w^b-\hat{\w}^{b}}\\
    &\leq \underbrace{\sum_{b=1}^K\EC{\langle\z^b,\hat{\w}^{b}-\w\rangle}}_{\mathtt{TERM}\text{-}\mathtt{A}} + \underbrace{\sum_{b=1}^KGL\EC{\Norm{\w^b-\hat{\w}^{b}}}}_{\mathtt{TERM}\text{-}\mathtt{B}}.
\end{align*}

We first give the bound  of $ \mathtt{TERM}\text{-}\mathtt{A}$. We define $\ell_1(\w)=\langle \z^1,\w\rangle+\frac{1}{\eta}\Norm{\w}^2$ and $\ell_b(\w)=\langle\z^b,\w\rangle$. By using Lemma \ref{lem:gh16}, we have 
\begin{equation}
\begin{aligned}
    &\sum_{b=1}^K\langle\z^b,\hat{\w}^{b+1}-\w\rangle+\frac{\Norm{\hat{\w}^2}^2-\Norm{\w}^2}{\eta}\leq\sum_{b=1}^K\ell_b(\hat{\w}^{b+1})-\ell_b(\w)\leq 0.
\end{aligned}    
\end{equation}

Therefore, we can derive
\begin{align*}
    \sum_{b=1}^K\langle\z^b,\hat{\w}^{b}-\w\rangle=&\sum_{b=1}^K\langle\z^b,\hat{\w}^{b}-\hat{\w}^{b+1}\rangle+\sum_{b=1}^K\langle\z^b,\hat{\w}^{b+1}-\w\rangle\\
    \leq&\sum_{b=1}^K\Norm{\z^b}\Norm{\hat{\w}^b-\hat{\w}^{b+1}}+\frac{\Norm{\hat{\w}^2}^2-\Norm{\w}^2}{\eta}\\
    \leq&\frac{\eta G^2 LT}{2}+\frac{D^2}{\eta},
\end{align*}
where the last inequality is due to $\Norm{\z^b}\leq LG$.

The key difference in the proof is the bound of the norm of the errors. We introduce the following lemma.
\begin{lem}\label{lem:2}
    Under the Assumptions \ref{gradient} and \ref{domain},  at any block $b$ of D-FTFCL, the norm of the errors is bounded by
    \begin{align*}
        \EC{\Norm{\e^{b+1}}^2}&\leq4e^2L^2G^2, \quad \EC{\Norm{\hat{\e}^{b+1}}^2}\leq120e^2L^2G^2.\\
    \end{align*}
\end{lem}
In the following, we explicitly characterize the impact of the delayed updates and can get
\begin{equation}\label{eq:15}
\begin{aligned}
    \Norm{\w^b-\hat{\w}^{b}}=&\Norm{\w^b-\y^{b}+\y^b-\tilde{\w}^b + \tilde{\w}^b-\hat{\w}^{b}}\\
   \leq & \Norm{\w^b-\y^{b}}+\Norm{\tilde{\w}^b-\y^{b}}+\Norm{\tilde{\w}^b-\hat{\w}^{b}}\\
    \leq & \frac{\eta}{2}\left(\Norm{\sum_{k=1}^{b-2}\s^k - \vv^k}+\Norm{\sum_{k=1}^{b-2}\vv^k-\z^k}+\Norm{\z^{b-2}+\z^{b-1}}\right)\\
   = & \frac{\eta}{2}\left(\Norm{\hat{\e}^{b-1}}+\Norm{\e^{b-1}}+\Norm{\z^{b-2}+\z^{b-1}}\right)\\
   \leq & \frac{\eta}{2}\left( 2eLG+\sqrt{120}eLG+2LG \right)\leq 20\eta LG,
\end{aligned}
\end{equation}
where the second inequality is due to Lemma \ref{lem:duchi11} and the last inequality is due to Lemma \ref{lem:2}.



By setting the learning rate $\eta = \frac{D}{G\sqrt{LT}}$, the final regret bound is 
\begin{align*}
    \EC{R_T} &\leq \sum_{b=1}^K\langle\EC{\z^b,\hat{\w}^{b}-\w\rangle} +GL\EC{\Norm{\w^b-\hat{\w}^{b}}}\\
    &\leq \frac{\eta G^2 LT}{2}+\frac{D^2}{\eta}+ 20\eta LTG^2\\
    &\leq O(DG\sqrt{LT})=O(DG\delta^{-1/2}\sqrt{T}).
\end{align*}
\subsection{Proof of Theorem \ref{thm:4}}
We first give some virtual decisions.
\begin{align*}
    \hat{\w}^{b+1}&=\arg\min_{\w\in\W}\left\langle\sum_{k=1}^b\z^k,\w\right\rangle+\frac{\mu L}{2}\sum_{k=1}^b\Norm{\w-\w^k}^2\\
    &=\arg\min_{\w\in\W}\left\langle\sum_{k=1}^b\z^k-\sum_{k=1}^b\mu L \w^k,\w\right\rangle+\frac{\mu bL}{2}\Norm{\w}^2.
\end{align*}

\begin{align*}
    \y^{b+1}&=\arg\min_{\y\in\W}\left\langle\sum_{k=1}^{b-2}\mathbf{v}^k,\y\right\rangle+\frac{\mu L}{2}\sum_{k=1}^b\Norm{\y-\w^k}^2\\
    &=\arg\min_{\y\in\W}\left\langle\sum_{k=1}^{b-2}\mathbf{v}^k-\sum_{k=1}^b\mu L \w^k,\y\right\rangle+\frac{\mu bL}{2}\Norm{\y}^2.
\end{align*}
\begin{align*}
    \tilde{\w}^{b+1}&=\arg\min_{\w\in\W}\left\langle\sum_{k=1}^{b-2}\z^k,\w\right\rangle+\frac{\mu L}{2}\sum_{k=1}^b\Norm{\w-\w^k}^2\\
    &=\arg\min_{\w\in\W}\left\langle\sum_{k=1}^{b-2}\mathbf{z}^k-\sum_{k=1}^b\mu L \w^k,\w\right\rangle+\frac{\mu bL}{2}\Norm{\w}^2.
\end{align*}

In the following, we begin to derive the regret bound
\begin{align*}
    \EC{R_T}&=\frac{1}{n}\sum_{t=1}^T\sum_{i=1}^n \EC{f_i^t(\w^t)-f_i^t(\w)}\\
    &\leq\frac{1}{n}\sum_{t=1}^T\sum_{i=1}^n\EC{\langle\g_i^t,\w^t-\w\rangle-\frac{\mu}{2}\Norm{\w^{t}-\w}}\\
    &=\sum_{b=1}^K\sum_{t=(b-1)L}^{bL}\E_{\C}\left[\langle\g^t,\w^b-\w\rangle-\frac{\mu}{2}\Norm{\w^{b}-\w}\right]\\
    &=\sum_{b=1}^K\E_{\C}\left[\langle\z^b,\w^b-\hat{\w}^{b}+\hat{\w}^{b}-\w\rangle-\frac{\mu L}{2}\Norm{\w^{b}-\w}\right]\\
    &=\sum_{b=1}^K\E_{\C}\left[\langle\z^b,\w^b-\hat{\w}^{b}\rangle\right]+\E\left[\langle\z^b,\hat{\w}^{b}-\w\rangle-\frac{\mu L}{2}\Norm{\w^{b}-\w}\right]\\
    &\leq \underbrace{\sum_{b=1}^K\E_{\C}\left[\langle\z^b,\hat{\w}^{b}-\w\rangle-\frac{\mu L}{2}\Norm{\w^{b}-\w}\right]}_{\mathtt{TERM}\text{-}\mathtt{A}} + \underbrace{\sum_{b=1}^KGL\EC{\Norm{\w^b-\hat{\w}^{b}}}}_{\mathtt{TERM}\text{-}\mathtt{B}},
\end{align*}
where the first inequality is due to Assumption~\ref{sconvex} and the last inequality is due to $\Norm{\z^b}\leq GL$.

We first give the bound of $\mathtt{TERM}\text{-}\mathtt{A}$. We define $\ell_b(\w)=\langle \z^b,\w\rangle+\frac{\mu L}{2}\Norm{\w-\w^b}^2$ and $\hat{\w}^{b+1}=\arg\min_{\w\in\W}\sum_{k=1}^b \ell_k(\w)$ in the strongly convex case. Then, we have 
\begin{equation}\label{eq:11}
\begin{aligned}
    &\sum_{b=1}^{K}\langle\z^b,\hat{\w}^{b}-\w\rangle-\frac{\mu L}{2}\Norm{\w^{b}-\w}\\
    &=\sum_{b=1}^{K}\langle\z^b,\hat{\w}^{b}-\hat{\w}^{b+1}+\hat{\w}^{b+1}-\w\rangle-\frac{\mu L}{2}\Norm{\w^{b}-\w}\\
    &\leq \sum_{b=1}^{K}\ell_b(\hat{\w}^{b+1})-\ell_b(\w)+GL\Norm{\hat{\w}^{b}-\hat{\w}^{b+1}}\\
    &\leq \sum_{b=1}^K GL\Norm{\hat{\w}^{b}-\hat{\w}^{b+1}}.
\end{aligned}
\end{equation}

Next, we give the bound of $GL\Norm{\hat{\w}^{b}-\hat{\w}^{b+1}}$. We define $F_b(\w)=\sum_{k=1}^b\ell_k(\w)$. It is not hard to verify that $F_b(\w)$ is $(bL\mu)$-strongly convex function and $\hat{\w}^{b+1}=\arg\min_{\w\in\W}F_b(\w)$.

By using Lemma \ref{lem:3}, for any $\x,\y\in\W$, we have
\begin{align*}
    |\ell_b(\x)-\ell_b(\y)|&\leq\langle\nabla\ell_b(\x),\x-\y\rangle\\
    &\leq \Norm{\nabla \ell_b(\x)}\Norm{\x-\y}\\
    &\leq (GL+\mu L D)\Norm{\x-\y}.
\end{align*}

Therefore, we can derive
\begin{align*}
    GL\Norm{\hat{\w}^{b+1}-\hat{\w}^b}^2\leq&\frac{2GL}{bL\mu}(F_b(\hat{\w}^b)-F_b(\hat{\w}^{b+1}))\\
    \leq&\frac{2GL}{bL\mu}(F_{b-1}(\hat{\w}^b)-F_{b-1}(\hat{\w}^{b+1})+\ell_b(\hat{\w}^b)-\ell_b(\hat{\w}^{b+1}))\\
    \leq &\frac{2GL}{bL\mu}(\ell_b(\hat{\w}^b)-\ell_b(\hat{\w}^{b+1}))\\
    \leq &\frac{2GL}{bL\mu}(GL+\mu DL)\Norm{\hat{\w}^{b+1}-\hat{\w}^b},
\end{align*}
which means
\begin{equation}\label{eq:12}
    GL\Norm{\hat{\w}^{b+1}-\hat{\w}^b}\leq \frac{2GL}{bL\mu}(GL+\mu DL).
\end{equation}

Moreover, according to (\ref{eq:15}), we have 
\begin{equation}\label{eq:115}
\begin{aligned}
    \Norm{\w^b-\hat{\w}^{b}}=&\Norm{\w^b-\y^{b}+\y^b-\tilde{\w}^b + \tilde{\w}^b-\hat{\w}^{b}}\\
   \leq & \Norm{\w^b-\y^{b}}+\Norm{\tilde{\w}^b-\y^{b}}+\Norm{\tilde{\w}^b-\hat{\w}^{b}}\\
   \leq & \frac{1}{\mu bL}\left(\Norm{\hat{\e}^{b-1}}+\Norm{\e^{b-1}}+\Norm{\z^{b-2}+\z^{b-1}}\right)\\
   \leq & \frac{1}{\mu bL}\left( 2eLG+\sqrt{120}eLG+2LG \right)\leq \frac{40LG}{\mu bL}.
\end{aligned}
\end{equation}


Therefore, by using Lemma \ref{lem:2}, (\ref{eq:11}) and (\ref{eq:12}), we have 
\begin{align*}
    \EC{R_T}&\leq \sum_{b=1}^K \frac{2GL}{bL\mu}(GL+\mu DL) + \frac{GL}{\mu b L}(40LG)\\
    &=L\sum_{b=1}^K \frac{2G}{b\mu}(G+\mu D) +\frac{40 G^2}{\mu b}\\
    &\leq O\left(L\frac{(G+\mu D)^2}{\mu}\log T\right),
\end{align*}
where the last inequality is due to $\sum_{b=1}^{T/L}\frac{1}{b}\leq \log (T/L)+1.$

\subsection{Proof of Theorem \ref{thm:5}}
We first recall the property of the  subgradient. For a subgradient $\g_i \in \partial f_i(\x)$ of a convex function $f_i(\cdot)$ at $\x$, we have $f_i(\y)\geq f_i(\x) + \langle\g_i,\y-\x \rangle$. Then, we first give the theoretical guarantee of the anytime online-to-batch conversion.
\begin{lem}\label{ashok:thm1}(Theorem 2 in \citet{cutkosky2019anytime}) We assume $\hat{\ell}^t(\x)$ is convex and satisfies $f(\x^t)-f(\x)\leq \E[\hat{\ell}^t(\x^t)-\hat{\ell}^t(\x)]$. Then for all $\x\in\W$ and $\x^T=\frac{\sum_{t=1}^T\alpha_t\w^t}{\alpha_{1:T}}$, the online-to-batch conversion guarantees:
    \begin{equation*}
        \E_{\C}[f(\x^T)-f(\x)]\leq \E_{\C}\left[\frac{\sum_{t=1}^T \hat{\ell}^t(\w^t) - \hat{\ell}^t(\x)}{\alpha_{1:T}}\right]
    \end{equation*}
\end{lem}

By setting $\alpha_t=1$, $\hat{\ell}^t_i(\w)=\langle \g_i^t,\w\rangle$, $\hat{\ell}^t(\w)=\langle \g^t,\w\rangle$ and $\g^t = \frac{1}{n}\sum_{i=1}^n \g_i^t$, we can obtain 
    \begin{equation*}
        \E_{\C}[f(\x^K)-f(\x^\star)]=\E_{\C}\left[\frac{1}{n}\sum_{i=1}^nf_i(\x^K)-f_i(\x^\star)\right]\leq \E_{\C}\left[\frac{R_K}{K}\right].
    \end{equation*}
We further have
\begin{align*}
    \EC{R_K} &= \sum_{t=1}^K\E_{\C}\left[\langle\g^t,\w^t-\w\rangle\right]\\
    &\leq \sum_{t=1}^K\langle\g^t,\w^t-\hat{\w}^{t}+\hat{\w}^{t}-\w\rangle\\
    &=\sum_{t=1}^K\langle \g^t,\hat{\w}^{t}-\w\rangle +\langle\g^t,\w^t-\hat{\w}^{t}\rangle\\
    &\leq \sum_{t=1}^K\langle\g^t,\hat{\w}^{t}-\w\rangle +G\Norm{\w^t-\hat{\w}^{t}}.
\end{align*}

Similar to the previous proof, we give the bound of the compression error.
\begin{lem}\label{lem:128-1}
    Under the Assumptions \ref{domain} and \ref{sgradient},  at any round $t$ of D-O2B with compression communication, the norm of the errors is bounded by
    \begin{align*}
       \EC{\Norm{\e^{t+1}}^2}&\leq4e^2\alpha_t^2G^2, \quad \EC{\Norm{\hat{\e}^{t+1}}^2}\leq 120e^2\alpha_t^2G^2.\\
    \end{align*}
\end{lem}

According to the previous proof of Theorem \ref{thm:1}, by using \eqref{eq:16},  Lemma \ref{lem:duchi11} and \ref{lem:128-1}, and setting $\alpha_t=1$, we can obtain the following
\begin{align*}
    \EC{R_K}&\leq \sum_{t=1}^K\langle\g^t,\hat{\w}^{t}-\w\rangle +G\Norm{\w^t-\hat{\w}^{t}}\\
    &\leq \frac{\eta G^2 K}{2}+\frac{D^2}{\eta}+ \frac{\eta}{2}\left(\Norm{\hat{\e}^{t-1}}+\Norm{\e^{t-1}}\right)\\
    &\leq  \frac{\eta G^2 K}{2}+\frac{D^2}{\eta}+\frac{13e\eta G}{2}\leq O(DG\sqrt{K}),
\end{align*}
where the last inequality due to $\eta = \frac{D}{G\sqrt{K}}$. 

Furthermore, we can obtain
\begin{equation*}
    \E_{\C}[f(\x^K)-f(\x^\star)]\leq \E\left[\frac{O(DG\sqrt{K})}{K}\right]= O\left(\frac{DG}{\sqrt{\delta T}}\right),
\end{equation*}
where the last equality is due to $T = K L$.

\subsection{Proof of Theorem \ref{thm:6}}
Following \citet{cutkosky2019anytime}, we set $\alpha_t=t$, $\hat{\ell}^t_i(\w)=\langle t\g^t_i,\w\rangle+\frac{\mu t}{2}\Norm{\w-\x^t}$, $\hat{\ell}^t(\w)=\frac{1}{n}\sum_{i=1}^n \hat{\ell}_i^t(\w)$, we can obtain 
    \begin{equation*}
        \E_{\C}[f(\x^K)-f(\x^\star)]\leq \EC{\frac{2R_K}{K(K+1)}}.
    \end{equation*}

In each round $t$, each learner updates its decision as follows:
\begin{align*}
    \w^{t+1}&=\arg\min_{\w\in\W}\left\langle\sum_{k=1}^t\s^k,\w\right\rangle+\frac{\mu }{2}\sum_{k=1}^tk\Norm{\w-\x^k}^2\\
    & = \arg\min_{\w\in\W}\left\langle\sum_{k=1}^t\s^k - \sum_{k=1}^t k\mu\x^k,\w\right\rangle+\frac{\mu t(t+1)}{4}\Norm{\w}^2,
\end{align*}
and we define two variables:
\begin{align*}
    \y^{t+1}&=\arg\min_{\y\in\W}\left\langle\sum_{k=1}^t\mathbf{v}^k,\y\right\rangle+\frac{\mu }{2}\sum_{k=1}^tk\Norm{\y-\x^k}^2\\
    &=\arg\min_{\y\in\W}\left\langle\sum_{k=1}^t\mathbf{v}^k- \sum_{k=1}^t k\mu\x^k,\y\right\rangle+\frac{\mu t(t+1)}{4}\Norm{\y}^2,
\end{align*}
\begin{align*}
\hat{\w}^{t+1}&=\arg\min_{\w\in\W}\left\langle\sum_{k=1}^tk\g^k,\w\right\rangle+\frac{\mu }{2}\sum_{k=1}^tk\Norm{\w-\x^k}^2\\
&=\arg\min_{\w\in\W}\left\langle\sum_{k=1}^tk\g^k- \sum_{k=1}^t k\mu\x^k,\w\right\rangle+\frac{\mu t(t+1)}{4}\Norm{\w}^2.
\end{align*}

Next, we give the regret bound on the loss function $\hat{\ell}^t(\w)$.
\begin{align*}
    R_K =& \sum_{t=1}^K \hat{\ell}^t(\w^{t}) - \hat{\ell}^t(\w)\\
    \leq&\sum_{t=1}^K \hat{\ell}^t(\hat{\w}^{t}) - \hat{\ell}^t(\w) + \hat{\ell}^t(\w^{t})- \hat{\ell}^t(\hat{\w}^{t}) \\
    \leq & \sum_{t=1}^K \hat{\ell}^t(\hat{\w}^{t}) - \hat{\ell}^t(\w) + (t G+\mu tD )\Norm{\hat{\w}^t-\w^{t}}.
\end{align*}

To bound the term $\sum_{t=1}^K \hat{\ell}^t(\hat{\w}^{t}) - \hat{\ell}^t(\w)$, we introduce a lemma.
\begin{lem}[Corollary 5 in \citet{cutkosky2019anytime}]\label{ashok:col5}
    Under the Assumptions \ref{sconvex}, \ref{domain} and \ref{sgradient}, by setting the weight $\alpha_t=t$, $\hat{\ell}^t(\w) = \langle \alpha_t\g^t, \w \rangle+\frac{\mu \alpha_t}{2}\Norm{\w-\x^t}^2$ and $\hat{\w}^{t+1} =\arg\min_{\w\in\W} \sum_{k=1}^t \hat{\ell}^t(\w)$, for any $\w \in \W$ anytime O2B can ensure 
    \begin{equation*}
        \sum_{t=1}^K \hat{\ell}^t(\hat{\w}^{t}) - \hat{\ell}^t(\w)\leq \frac{K(\mu D + G)^2}{\mu}.
    \end{equation*}
\end{lem}

Then we bound the term $\Norm{\w^t-\hat{\w}^t}$. By using Lemma \ref{lem:duchi11} and \ref{lem:128-1}, and setting $\alpha_t=t$, we have  
\begin{equation}\label{eq:17}
\begin{aligned}
    \EC{\Norm{\w^t-\hat{\w}^{t}}}&\leq\frac{2}{\mu t(t+1)}\left(\EC{\Norm{\hat{\e}^{t-1}}}+\EC{\Norm{\e^{t-1}}}\right)\\
    &\leq \frac{30eG}{\mu (t+1)}.
\end{aligned}
\end{equation}

Finally, by using Lemma \ref{ashok:col5} and \eqref{eq:17}, we can derive the regret bound
\begin{align*}
    \EC{R_K}&\leq  \frac{K(\mu D + G)^2}{\mu}+\sum_{t=1}^K \frac{30 (G+\mu D)^2 et}{\mu t}\\
    &\leq \frac{2K(G+\mu D)^2}{\mu}+\frac{30 e K  (G+\mu D)^2}{\mu}=O\left(\frac{K(G+\mu D)^2}{\mu}\right).
\end{align*}

The final convergence rate is 
    \begin{equation*}
        \E_{\C}[f(\x^K)-f(\x^\star)] \leq \EC{\frac{2R_K}{K(K+1)}} \leq O\left(\frac{(G+\mu D)^2}{\mu K}\right) = O\left(\frac{(G+\mu D)^2}{\delta \mu T}\right).
    \end{equation*}

\textbf{Additional discussion.} It is worth noting that our method involves only $K$ updates, with the total communication rounds amounting to $T = \lceil \frac{1}{\delta}\rceil K$. In contrast, standard compression-based methods (e.g., O2B with Algorithm \ref{alg:1} and standard compressor) typically perform $T$ updates over $T$ communication rounds, yielding a convergence rate for convex loss functions of:
\begin{equation*}
    \E[f(\x^T)-f(\x^\star)]\leq \E\left[\frac{R_T}{T}\right]=\frac{DG}{\delta\sqrt{T}}.
\end{equation*}

Compared to these methods, our approach achieves not only a faster convergence rate but also lower sample complexity under the same communication budget (i.e., the same number of communication rounds $T$). From a high-level perspective, although our algorithm performs fewer updates, the information incorporated in each update exhibits significantly lower bias compared to standard compression, thereby resulting in faster convergence.

Although using Algorithm \ref{alg:2} achieves the same convergence rate, it incurs redundant gradient queries.

\subsection{Proof of the Unidirectional Compression}\label{appendix:unidierectional}
It is worth noting that under the unidirectional compression, learner $i$ updates the decision according to
\begin{equation*}
    \w^{t+1} = \y^{t+1}=\arg\min_{\y\in\W}\left\langle\sum_{k=1}^t\mathbf{v}^k,\y\right\rangle+\frac{1}{\eta}\Norm{\y}^2.
\end{equation*}

Therefore, we can obtain the regret bound 
\begin{align*}
        \EC{R_T}&=\EC{\frac{1}{n}\sum_{t=1}^T\sum_{i=1}^nf_i^t(\w^t)-\frac{1}{n}\sum_{t=1}^T\sum_{i=1}^nf_i^t(\w)}\\
    &\leq \sum_{t=1}^T\EC{\langle\g^t,\hat{\w}^{t}-\w \rangle} +G\EC{\Norm{\y^t-\hat{\w}^{t}}}.
\end{align*}

According to the proof of Theorem \ref{thm:1}, we have 
\begin{equation*}
\EC{\Norm{\y^t-\hat{\w}^{t}}}\leq\frac{\eta}{2}\EC{\Norm{\e^t}}\leq \frac{\eta G}{\delta}.
\end{equation*}

By using \eqref{eq:16} and setting $\eta = \frac{D\sqrt{\delta}}{G}$, the final regret bound is 
\begin{align*}
        \EC{R_T}&=\EC{\frac{1}{n}\sum_{t=1}^T\sum_{i=1}^nf_i^t(\w^t)-\frac{1}{n}\sum_{t=1}^T\sum_{i=1}^nf_i^t(\w)}\\
    &\leq\frac{\eta G^2T}{2} + \frac{D^2}{\eta} + \frac{\eta G}{\delta} \leq O(DG\sqrt{\delta^{-1} T}).
\end{align*}

Notably, the regret bound of $O(\delta^{-1}\log T)$ for strongly convex loss functions can be derived using an analogous proof under unidirectional compression.

\section{Proof of Supporting Lemmas}
\subsection{Proof of Lemma \ref{lem:1}}

We first give the bound of the compression error of each learner. Following the proof in \citet{karimireddy2019error}, we have
\begin{align*}
\mathbb{E}_{\C}\left[\|\e^{t+1}\|^2\right]&= \mathbb{E}_{\C}\!\left[\left\|\frac{1}{n}\sum_{i=1}^n \e_i^{t+1}\right\|^2\right]
   \le \frac{1}{n}\sum_{i=1}^n \mathbb{E}_{\C}\!\left[\|\e_i^{t+1}\|^2\right]
   \\
   &= \frac{1}{n}\sum_{i=1}^n \mathbb{E}_{\C}\!\left[\|\e_i^{t} + \g_i^{t} - \mathcal{C}(\e_i^{t}+\g_i^{t})\|^2\right] \\
&\le \frac{1-\delta}{n}\sum_{i=1}^n \mathbb{E}_{\C}\left[\|\e_i^{t}+\g_i^{t}\|^2\right] \\
&\le (1-\delta)(1+\alpha_1)\frac{1}{n}\sum_{i=1}^n \mathbb{E}_{\C}\!\left[\|\e_i^{t}\|^2\right]
   + (1-\delta)(1+\alpha_1^{-1})\frac{1}{n}\sum_{i=1}^n\Norm{\g_i^t}^2 \\
&\le  (1-\delta)(1+\alpha_1)\frac{1}{n}\sum_{i=1}^n \mathbb{E}_{\C}\!\left[\|\e_i^{t}\|^2\right]
   + (1-\delta)(1+\alpha_1^{-1})G^2,
\end{align*}
where the third inequality is due to $\Norm{\x+\y}^2\leq (1+\alpha_1)\Norm{\x}^2+(1+\alpha_1^{-1})\Norm{\y}^2$.  By  setting $\alpha_1 = \frac{\delta}{2-2\delta}$ and summing up, we can derive
\begin{equation}
\begin{aligned}
    \mathbb{E}_{\C}\left[\|\e^{t+1}\|^2\right]\leq&\frac{1}{n}\sum_{i=1}^n \mathbb{E}_{\C}\left[\|\e_i^{t+1}\|^2\right]\leq\sum_{k=0}^{t-1}\left((1-\delta)(1+\alpha_1)\right)^{t-k}(1-\delta)(1+\alpha_1^{-1})G^2\\
    \leq&\frac{(1-\delta)(1+\alpha_1^{-1})G^2}{1-(1-\delta)(1+\alpha_1)}=\frac{(1-\delta)(1+\alpha_1^{-1})G^2}{\delta-\alpha_1+\alpha_1\delta}\\
    \leq&\frac{4(1-\delta)G^2}{\delta^2}.
\end{aligned}    
\end{equation}

As for the upper bound of $\EC{\Norm{\hat{\e}^{t+1}}^2}$, we have the following
\begin{equation}\label{eq:32}
  \begin{aligned}
\EC{\Norm{\hat{\e}^{t+1}}^2}&=\EC{\Norm{\sum_{k=1}^{t}\vv^k-\s^{k}}^2}=\EC{\Norm{\sum_{k=1}^{t}\mathbf{v}^k-\sum_{k=1}^{t-1}\s^k-\C\left(\hat{\e}^t+\vv^t\right)}^2}\\
& =\EC{\Norm{\mathbf{v}^t+\hat{\e}^t-\C\left(\hat{\e}^t+\vv^t\right)}^2}\\
    &\leq (1-\delta)\EC{\Norm{\vv^{t}+\hat{\e}^t}}\\
    &\leq (1-\delta)(1+\alpha_2)\EC{\Norm{\hat{\e}^{t}}^2}+(1-\delta)(1+\alpha_2^{-1})\EC{\Norm{\mathbf{v}^{t}}^2}.
 \end{aligned}  
\end{equation}

As for the second term $\EC{\Norm{\vv^{t}}^2}$, we have
\begin{equation}\label{eq:31}
  \begin{aligned}
          \EC{\Norm{\vv^{t}}^2}&\leq\frac{1}{n}\sum_{i=1}^n\EC{\Norm{\vv^{t}_i}^2}=\frac{1}{n}\sum_{i=1}^n\EC{\Norm{\vv^{t}_i-(\e_i^{t}+\g_i^{t})+(\e_i^{t}+\g_i^{t})}^2}\\
    &\leq \frac{2}{n}\sum_{i=1}^n\EC{\Norm{\C(\e_i^{t}+\g_i^{t}))-(\e_i^{t}+\g_i^{t})}^2+\frac{2}{n}\sum_{i=1}^n\Norm{\e_i^{t}+\g_i^{t}}^2}\\
    &\leq \frac{2(1-\delta)}{n}\sum_{i=1}^n\EC{\Norm{\e_i^{t}+\g_i^{t}}^2}+\EC{\frac{2}{n}\sum_{i=1}^n\Norm{\e_i^{t}+\g_i^{t}}^2}\\
    &= \frac{2(2-\delta)}{n}\sum_{i=1}^n\EC{\Norm{\e_i^{t}+\g_i^{t}}^2}\\
    &\leq \frac{4(2-\delta)}{n}\sum_{i=1}^n\left(\EC{\Norm{\e_i^{t}}^2}+\Norm{\g_i^{t}}^2\right)\\
    &\leq 4(2-\delta)\left(\frac{4(1-\delta)G^2}{\delta^2}+G^2\right)\\
    &\leq 8\left(\frac{4G^2}{\delta^2}+\frac{1}{\delta^2}G^2\right)\leq \frac{40G^2}{\delta^2}.
  \end{aligned}  
\end{equation}

By setting $\alpha_2=\frac{\delta}{2-2\delta}$ and plugging \eqref{eq:31} into \eqref{eq:32}, we can derive
\begin{align*}
    \EC{\Norm{\hat{\e}^{t+1}}^2}\leq& (1-\delta)(1+\alpha_2)\EC{\Norm{\hat{\e}^{t}}^2}+(1-\delta)(1+\alpha_2^{-1})\frac{40G^2}{\delta^2}\\
    \leq&\sum_{k=0}^{t-1}\left((1-\delta)(1+\alpha_2)\right)^{t-k}(1-\delta)(1+\alpha_2^{-1})\frac{40G^2}{\delta^2}\\
    \leq &\frac{(1-\delta)(1+\alpha_2^{-1})\frac{40G^2}{\delta^2}}{1-(1-\delta)(1+\alpha_2)}\leq \frac{160(1-\delta)G^2}{\delta^4}.
\end{align*}
\subsection{Proof of Lemma \ref{lem:2}}


According to Lemma \ref{rpc}, by choosing the compression round $L=\lceil1/\delta\rceil$, we can ensure $\mathbb{E}_{\C}\left[\Norm{\mathbf{r}^{L+1}-\x}^2\right]\leq \frac{1}{e}\Norm{\x}^2$, where $\mathbf{r}^{L+1}$ is the output of Algorithm \ref{rpc}. Therefore, we have 
\begin{equation}\label{eq:33}
    \EC{\Norm{\e_i^{b-1} + \z_i^{b-1}-\vv_i^{b-1}}^2}=\EC{\Norm{\e_i^{b-1} + \z_i^{b-1}-\text{FCC}(\e_i^{b-1}+\z_i^{b-1}, \C(\cdot), L)}^2}\leq \frac{1}{e}\Norm{\e_i^{b-1} + \z_i^{b-1}}^2.
\end{equation}
Therefore, we can derive
\begin{align*}
\mathbb{E}_{\C}\left[\|\e^{b+1}\|^2\right]
&= \mathbb{E}_{\C}\!\left[\left\|\frac{1}{n}\sum_{i=1}^n \e_i^{b+1}\right\|^2\right]\le \frac{1}{n}\sum_{i=1}^n \mathbb{E}_{\C}\!\left[\|\e_i^{b+1}\|^2\right]\\
   &= \frac{1}{n}\sum_{i=1}^n \mathbb{E}_{\C}\!\left[\|\e_i^{b} + \z_i^{b}- \vv_i^{b-1}\|^2\right] \\
&\le \frac{1}{en}\sum_{i=1}^n \mathbb{E}_{\C}\left[\|\e_i^{b}+\z_i^{b}\|^2\right] \\
&\le (1+\alpha_1)\frac{1}{e n}\sum_{i=1}^n \mathbb{E}_{\C}\!\left[\|\e_i^{b}\|^2\right]
   + (1+\alpha_1^{-1})\frac{1}{en}\sum_{i=1}^n\Norm{\z_i^{b}}^2 \\
&\le  (1+\alpha_1)\frac{1}{en}\sum_{i=1}^n \mathbb{E}_{\C}\!\left[\|\e_i^{b}\|^2\right]
   + \frac{1+\alpha_1^{-1}}{e}L^2G^2,
\end{align*}
where the second equality is due to \eqref{eq:33}. By  setting $\alpha_1 = \frac{e-1}{2}$ and summing up, we can derive
\begin{align*}
    \frac{1}{n}\sum_{i=1}^n \mathbb{E}_{\C}\left[\|\e_i^{b+1}\|^2\right]\leq&\sum_{k=0}^{b-1}\left(\frac{1+\alpha_1}{e}\right)^{b-k}(1+\alpha_1^{-1})\frac{L^2G^2}{e}\\
    \leq&\frac{(1+\alpha_1^{-1})L^2G^2}{e-(1+\alpha_1)} \leq\frac{4L^2G^2}{e(1-1/e)^2}\leq 4e^2L^2G^2.
\end{align*}

As for the term $\EC{\Norm{\hat{\e}^{b+1}}^2}$, we have the following
\begin{equation}\label{eq:34}
    \begin{aligned}
        \EC{\Norm{\hat{\e}^{b+1}}^2} &= \EC{\Norm{\hat{\e}^b + \vv^b-\text{FCC}\left(\hat{\e}^b + \vv^b, \C(\cdot), L\right)}^2}\\
    &\leq \frac{1}{e}\EC{\Norm{\hat{\e}^b + \vv^b}^2}\\
    &\leq \frac{1+\alpha_2}{e}\EC{\Norm{\hat{\e}^{b}}^2}+\frac{1+\alpha_2^{-1}}{e}\EC{\Norm{\mathbf{v}^{b}}^2},
    \end{aligned}
\end{equation}
where the first equality is due to \eqref{eq:33}. To bound the term $\EC{\Norm{\vv^{b+1}}^2}$, we have
\begin{equation}\label{eq:35}
    \begin{aligned}
    \EC{\Norm{\vv^{b}}^2}&\leq\frac{1}{n}\sum_{i=1}^n\EC{\Norm{\vv^{b}_i}^2}\\
    &=\frac{1}{n}\sum_{i=1}^n\EC{\Norm{\vv^{b}_i-(\e_i^{b}+\z_i^{b})+(\e_i^{b}+\z_i^{b})}^2}\\
    &\leq \frac{2}{n}\sum_{i=1}^n\EC{\Norm{\vv^{b}_i-(\e_i^{b}+\z_i^{b})}^2}+\EC{\frac{2}{n}\sum_{i=1}^n\Norm{\e_i^{b}+\z_i^{b}}^2}\\
    &\leq \frac{2/e}{n}\sum_{i=1}^n\EC{\Norm{\e_i^{b}+\z_i^{b}}^2}+\EC{\frac{2}{n}\sum_{i=1}^n\Norm{\e_i^{b}+\z_i^{b}}^2}\\
    &= \frac{2+2/e}{n}\sum_{i=1}^n\EC{\Norm{\e_i^{b}+\z_i^{b}}^2}\\
    &\leq \frac{4+4/e}{n}\sum_{i=1}^n\left(\EC{\Norm{\e_i^{b}}^2}+\Norm{\z_i^{b}}^2\right)\\
    &\leq (4+4/e)\left(4e^2L^2G^2 +L^2G^2\right)= 30e^2L^2G^2.        
    \end{aligned}
\end{equation}
By setting $\alpha_2=\frac{e-1}{2}$ and plugging \eqref{eq:35} into \eqref{eq:34}, we can derive
\begin{align*}
    \EC{\Norm{\hat{\e}^{b+1}}^2}\leq& \frac{1}{e}(1+\alpha_2)\EC{\Norm{\hat{\e}^{b}}^2}+\frac{1+\alpha_2^{-1}}{e}\EC{\Norm{\vv^{b}}^2}\\
    \leq& \frac{1}{e}(1+\alpha_2)\EC{\Norm{\hat{\e}^{b}}^2}+\frac{1+\alpha_2^{-1}}{e}30e^2L^2G^2\\
    \leq&\sum_{k=0}^{t-1}\left(\frac{1}{e}(1+\alpha_2)\right)^{t-k}\frac{1+\alpha_2^{-1}}{e}30e^2L^2G^2\\
    \leq &\frac{(1+\alpha_2^{-1})30e^2L^2G^2}{e-(1+\alpha_2)}\leq 120e^2L^2G^2,
\end{align*}
where the last inequality is due to $\frac{1+e}{(e-1)^2}\leq2$.

\subsection{Proof of Lemma \ref{lem:128-1}}

We first give the bound of the compression error of each learner. Following the proof in \citet{karimireddy2019error}, we have
\begin{align*}
\mathbb{E}_{\C}\left[\|\e^{t+1}\|^2\right]&\le \frac{1}{n}\sum_{i=1}^n \mathbb{E}_{\C}\!\left[\|\e_i^{t+1}\|^2\right]
   \\
   &= \frac{1}{n}\sum_{i=1}^n \mathbb{E}_{\C}\!\left[\|\e_i^{t} + \alpha_t\g_i^{t} - \text{FCC}(\e_i^{t}+ \alpha_t\g_i^{t}, \C(\cdot), L)\|^2\right] \\
&\le \frac{1}{en}\sum_{i=1}^n \mathbb{E}_{\C}\left[\|\e_i^{t}+ \alpha_t\g_i^{t}\|^2\right] \\
&\le (1+\beta_1)\frac{1}{en}\sum_{i=1}^n \mathbb{E}_{\C}\!\left[\|\e_i^{t}\|^2\right]
   + (1+\beta_1^{-1})\frac{1}{en}\sum_{i=1}^n\Norm{ \alpha_t\g_i^t}^2 \\
&\le  (1+\beta_1)\frac{1}{en}\sum_{i=1}^n \mathbb{E}_{\C}\!\left[\|\e_i^{t}\|^2\right]
   + (1+\beta_1^{-1})\alpha_t^2 G^2.
\end{align*}

By  setting $\beta_1 = \frac{e-1}{2}$ and summing up, we can derive
\begin{equation}\label{128-1}
\begin{aligned}
    \mathbb{E}_{\C}\left[\|\e^{t+1}\|^2\right]\leq&\frac{1}{n}\sum_{i=1}^n \mathbb{E}_{\C}\left[\|\e_i^{t+1}\|^2\right]\leq \alpha_{t}^2 \sum_{k=0}^{t-1}\left(\frac{1}{e}(1+\beta_1)\right)^{t-k}\frac{1}{e}(1+\beta^{-1}) G^2\\
    \leq&\frac{\alpha_{t}^2(1+\beta^{-1})G^2}{e-(1+\beta_1)}\leq4e^2\alpha_t^2G^2,
\end{aligned}    
\end{equation}
where the third inequality is due to $\alpha_{t}\geq \alpha_{t-1}$.

As for the upper bound of $\EC{\Norm{\hat{\e}^{t+1}}^2}$, we similar inequality
\begin{equation}\label{128-32}
  \begin{aligned}
\EC{\Norm{\hat{\e}^{t+1}}^2}&=\EC{\Norm{\sum_{k=1}^{t}\vv^k-\s^{k}}^2}=\EC{\Norm{\sum_{k=1}^{t}\mathbf{v}^k-\sum_{k=1}^{t}\s^k-\text{FCC}\left(\sum_{k=1}^{t}\mathbf{v}^k-\sum_{k=1}^{t}\s^k,\C(\cdot),L\right)}^2}\\
    &\leq \frac{1}{e}\EC{\Norm{\vv^{t}+\hat{\e}^t}}\leq \frac{1}{e}(1+\beta_2)\EC{\Norm{\hat{\e}^{t}}^2}+ \frac{1}{e}(1+\beta_2^{-1})\EC{\Norm{\mathbf{v}^{t}}^2}.
 \end{aligned}  
\end{equation}

As for the second term $\EC{\Norm{\vv^{t}}^2}$, we have
\begin{equation}\label{128-31}
  \begin{aligned}
          \EC{\Norm{\vv^{t}}^2} &\leq \frac{1}{n}\sum_{i=1}^n\EC{\Norm{\vv^{t}_i-(\e_i^{t}+\alpha_t \g_i^{t})+(\e_i^{t}+\alpha_t\g_i^{t})}^2}\\
    &\leq \frac{2}{n}\sum_{i=1}^n\EC{\Norm{\vv^{t}_i-(\e_i^{t}+\alpha_t\g_i^{t})}^2+\frac{2}{n}\sum_{i=1}^n\Norm{\e_i^{t}+\alpha_t\g_i^{t}}^2}\\
    &\leq \frac{2/e}{n}\sum_{i=1}^n\EC{\Norm{\e_i^{t}+\alpha_t\g_i^{t}}^2}+\EC{\frac{2}{n}\sum_{i=1}^n\Norm{\e_i^{t}+\alpha_t\g_i^{t}}^2}\\
    &= \frac{2+2/e}{n}\sum_{i=1}^n\EC{\Norm{\e_i^{t}+\alpha_t\g_i^{t}}^2}\\
    &\leq \frac{4+4/e}{n}\sum_{i=1}^n\left(\EC{\Norm{\e_i^{t}}^2}+\Norm{\alpha_t\g_i^{t}}^2\right)\\
    &\leq (4+4/e)\left(4e^2\alpha_{t-1}^2G^2+\alpha_t^2G^2\right)\\
    &\leq (4+4/e)\left(4e^2\alpha_t^2G+\alpha_t^2G^2\right)\leq30e^2\alpha_t^2G^2.
  \end{aligned}  
\end{equation}

By setting $\beta_2=\frac{e-1}{2}$ and plugging \eqref{128-31} into \eqref{128-32}, we can derive
\begin{align*}
    \EC{\Norm{\hat{\e}^{t+1}}^2} \leq& \frac{1}{e}(1+\beta_2)\EC{\Norm{\hat{\e}^{t}}^2}+\frac{1}{e}(1+\beta_2^{-1})30e^2\alpha_t^2G^2\\
    \leq&\alpha_t^2\sum_{k=0}^{t-1}\left(\frac{1}{e}(1+\beta_2)\right)^{t-k}\frac{1}{e}(1+\beta_2^{-1})30e^2G^2\\
    \leq &\frac{(1+\beta_2^{-1})30e^2\alpha_t^2G^2}{e-(1+\beta_2)} \leq 120e^2\alpha_t^2G^2.
\end{align*}



\end{document}